%% file: main.tex
\documentclass{article}

\usepackage[final]{corl_2026} 

\usepackage{graphicx}
\usepackage{booktabs}
\usepackage{multirow}
\usepackage{xcolor}
\usepackage{amsmath,amssymb}
\usepackage{subcaption}
\usepackage{algorithm}
\usepackage{algorithmic}
\usepackage{xspace}
\usepackage{enumitem}
\usepackage{wrapfig}
\usepackage{makecell}
\usepackage{bbm}
\usepackage{tabularx}
\usepackage{color,colortbl}
\usepackage{hyperref}
\usepackage{cleveref}

\usepackage{amsmath}
\usepackage{amssymb}
\usepackage{mathtools}
\usepackage{amsthm}
\usepackage{mwe}
\usepackage{pifont}
\usepackage{wrapfig}
\usepackage{amssymb}
\usepackage{colortbl}
\usepackage{multirow}
\usepackage{multicol}
\usepackage{dsfont}

\usepackage{makecell}
\usepackage{bbm}

\usepackage[most]{tcolorbox}

\newcommand{\eg}{\textit{e}.\textit{g}.}
\newcommand{\ie}{\textit{i}.\textit{e}.}

\newcommand\blfootnote[1]{
        \begingroup
        \renewcommand\thefootnote{}\footnote{#1}
        \addtocounter{footnote}{-1}
        \endgroup
    }
\newcommand{\method}{PROSE\xspace}
\newcommand{\mainsec}[1]{\textcolor{red}{#1}}

\usepackage{titlesec}

\title{PROSE: Training-Free Egocentric Scene Registration with Vision-Language Models}

\author{
  \textbf{Zhiang Chen$^{*\,1}$ \quad Nahyuk Lee$^{*}$ \quad Boyang Sun$^{1}$ \quad Taein Kwon$^{2}$} \\
  \textbf{Marc Pollefeys$^{1}$ \quad Zuria Bauer$^{\dagger\,1}$ \quad Sunghwan Hong$^{\dagger\,\ddagger,1,3}$} \\[2pt]
  \normalfont\normalsize
  $^{1}$ETH Zurich \quad $^{2}$VGG, University of Oxford \quad $^{3}$ETH AI Center\\
  \href{https://RCKola.github.io/prose}{\texttt{https://RCKola.github.io/prose}}
}

\begin{document}
\maketitle
\vspace{-1.5cm}
\blfootnote{$^*$Equal contribution. $^\dagger$Equal supervision.}
\blfootnote{$^\ddagger$Corresponding author.}
\input{corl26/sections/0_abstract}

\input{corl26/sections/1_introduction}
\input{corl26/sections/2_related_works}
\input{corl26/sections/3_method}
\input{corl26/sections/4_experiments}

\input{corl26/sections/5_conclusion}


\clearpage

\bibliography{corl26/references}  

\input{corl26/sections/suppl}

\end{document}

%% file: corl26/sections/0_abstract.tex
\input{corl26/figures/1_teaser}
\vspace{-1mm}
\begin{abstract}
Registering two captures of the same indoor space taken at different times underpins persistent spatial memory for robots and AR systems, yet the realistic version of this task is egocentric and its most scalable form is RGB-only. Head-mounted cameras yield blurry, fast-moving, partially overlapping views from which dense geometry is hard to recover. Classical registration leans on exactly the clean point clouds this setting lacks, while learned scene-graph methods require a pre-built or annotated graph and a trained matcher that we find brittle under egocentric data. We take a different route, using a pretrained vision-language model as the source of both scene understanding and cross-scan matching. Our method, \textbf{\method} (\textbf{Pro}mpted \textbf{S}cene r\textbf{E}gistration), lifts each RGB sequence into an object-level 3D scene graph using off-the-shelf foundation models for geometry, segmentation, and language, then prompts the same VLM to match object instances across the two RGB sequences. To make this matching tractable and reliable, we leverage object heights as a prior and verify each proposed match with a paired same/different query, then solve for the rigid transform by hypothesizing a candidate per matched object and selecting the one with the strongest geometric consensus. \method adds no learned parameters and requires no depth sensor, training, or annotated graph. On the egocentric Aria Digital Twin and Aria Everyday Activities benchmarks, it outperforms both geometric and learned scene-graph baselines in registration accuracy, on ground-truth and RGB-reconstructed point clouds alike, and the scene graph it produces transfers directly to downstream tasks.

\vspace{-1mm}
\end{abstract}
\keywords{Scene Registration, Vision-Language Models, Egocentric Perception, Scene Graphs, Training-Free, 3D Scene Understanding}

%% file: corl26/figures/1_teaser.tex
\begin{figure}[h]
    \centering
    \includegraphics[width=\linewidth]{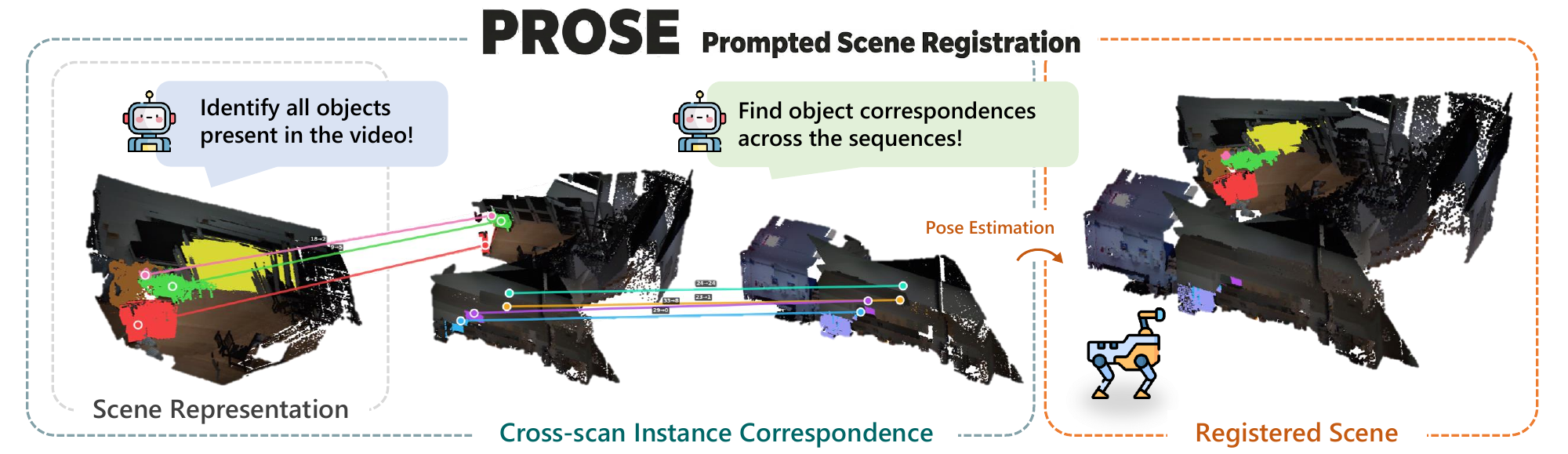}
   \caption{\textbf{Overview of \method.} From two egocentric RGB sequences of the same scene captured at different times with partial overlap, \method prompts a pretrained VLM to (i) list the objects in each sequence, lifting them into an object-level 3D scene graph, and (ii) match instances across the two scans. The resulting correspondences drive a training-free pose estimation that registers the two scenes. No depth sensor, training, or annotated graph is required.}
   \vspace{-10pt}
    \label{fig:teaser}
\end{figure}

%% file: corl26/sections/1_introduction.tex
\section{Introduction}
\label{sec:intro}
\vspace{-6.5pt}
Aligning two captures of the same indoor scene taken at different times is a primitive for persistent autonomy: a robot revisiting a room it mapped last month, an AR headset re-anchoring content across sessions, or a fleet of agents sharing a map all reduce to recovering a rigid transform between two RGB sequences with moved furniture and partial overlap~\cite{11024207,han2026geometric}. The deployment-realistic version is egocentric~\cite{sun2026lime,zhang2026chronograph,yang2026map}, where head-mounted cameras introduce motion blur, fast viewpoint changes, and rearranged objects that degrade the geometric signal point-cloud registration relies on. The landmark objects in the scene, however, remain recognizable, and they are also what an embodied agent ultimately cares about: a robot acting in the aligned scene needs to know what populates it and how it is arranged, not just where the cameras stood.

Existing methods provide neither robustly. The two families that dominate the cross-time scene registration literature both come up short in the egocentric RGB setting.
Pure geometric methods~\cite{yang2020teaser,rusu2009fast,FCGF2019, Seo_BUFFERX_arXiv_2025} discard semantics and match points across the entire target, where noisy egocentric RGB reconstructions, built from monocular depth under narrow field-of-view, irregular viewpoint coverage, and frequent motion blur, leave too few reliable matches to recover the transform.
Learned scene-graph registration~\cite{sarkar2023sgaligner,xie2024sg,11024207} restores the semantic prior, but assumes the scene graph is given, either as ground-truth annotation or from a separately trained front-end, and learns a bespoke matcher that, in our evaluation, does not transfer to the egocentric distribution. In both cases, scene understanding is either discarded or assumed, never produced as part of solving the alignment.

This points to a different prior. Geometry may be unreliable in egocentric capture, but objects remain recognizable across time. Our central observation is that, cross-time instance matching can be cast as a visual comparison task that pretrained VLMs already handle well. Recent work has applied VLMs to single-scene 3D understanding~\cite{qi2025gpt4scene,gom2026aaai}; we extend this to the cross-time setting. Prompted appropriately, the same VLM that identifies the objects can also match them across two captures of it. Object understanding and cross-scan matching therefore draw on VLMs as the central engine, and the scene graph built for matching doubles as a reusable scene representation.

We realize this in \textbf{Pro}mpted \textbf{S}cene r\textbf{E}gistration, namely \textbf{PROSE}, a training-free pipeline for cross-time egocentric scene alignment. PROSE first lifts RGB into per-subscan point clouds with a geometric foundation model~\cite{wang2026vggt}, prompts a pretrained VLM~\cite{qwen3.6-27b,achiam2023gpt,yoon2025visual} to list the objects in each scene, and uses SAM3~\cite{carion2025sam} to turn those names into temporally consistent instance masks, which a fusion step consolidates into a per-subscan scene graph (Fig.~\ref{fig:teaser}). The same VLM is then asked to match instances across the two scans. Naive prompting fails here, the query is intractably large, confusable objects at different heights compete, and the VLM frequently hallucinates matches, so we introduce a small set of prompting and filtering steps to make the matching reliable, then estimate the transform from the resulting correspondences via per-instance matching and RANSAC~\cite{fischler1981random}.

We evaluate PROSE on the egocentric Aria Digital Twin~\cite{pan2023aria} and Aria Everyday Activities~\cite{lv2024aria} datasets, which capture diverse indoor scenes under realistic motion and scene change. PROSE improves over both geometric and learned scene-graph baselines on registration recall, rotation, and translation, on both ground-truth and RGB-reconstructed point clouds, and ablations isolate the contribution of each stage and each design choice within the pipeline. PROSE also emits an open-vocabulary scene graph alongside the transform, which a simulated agent uses to plan paths over a registered map. Our contributions are:
\begin{itemize}
    \item We introduce \textbf{PROSE}, a training-free, RGB-only pipeline for cross-time egocentric scene registration that jointly recovers cross-scan instance correspondences and a rigid transform while producing an open-vocabulary scene graph in the same pass.
    \item We show that a pretrained VLM, when prompted appropriately, can serve as a cross-scan instance-correspondence engine: height-binned Set-of-Marks prompting with a verification pass is sufficient to drive correspondence without any training.
    \item We show that on egocentric datasets~\cite{lv2024aria,pan2023aria}, PROSE substantially improves over both geometric and learned scene-graph baselines on registration, and that the scene graphs it produces are reliable to support downstream use.
\end{itemize}

%% file: corl26/sections/2_related_works.tex
\section{Related Work}
\label{sec:related}

\noindent\textbf{Scene graph registration.} Cross-time alignment of indoor scenes via scene graphs originates with SGAligner~\cite{sarkar2023sgaligner}, which learns a multi-modal embedding over node labels, points, and structural relations and matches across two pre-segmented graphs. SG-PGM~\cite{xie2024sg} reformulates the task as partial graph matching with a GNN, reusing GeoTransformer point features and adding a learnable top-$k$ matching layer. SG-Reg~\cite{11024207} targets cross-distribution generalization by fusing BERT label features, GNN topology, and PointNet shape, and pairs the matcher with a robust pose back-end, though its front-end still assumes a fused metric point cloud with semantic labels. All three are learned on a target distribution, presuppose curated 3D inputs, and decouple scene graph construction from matching. PROSE differs on every axis: training-free, RGB-only, and producing the scene graph through the same VLM that drives correspondence.

\noindent\textbf{Geometric and learned 3D registration.} Classical correspondence-based registration pairs a local descriptor with a robust estimator: TEASER++~\cite{yang2020teaser} accepts up to $99\%$ outliers via a truncated least-squares relaxation and is the standard back-end for FPFH~\cite{rusu2009fast} and FCGF~\cite{FCGF2019} descriptor matches. GeoTransformer~\cite{qin2022geometric} replaces the descriptor-matching pipeline with a coarse-to-fine attention model over superpoints and is the de facto learned baseline for indoor benchmarks. BUFFER-X~\cite{Seo_BUFFERX_arXiv_2025} pushes toward zero-shot generality with patch-level coordinate normalization and distribution-aware downsampling. While specialized in the registration task, none of these methods exploit semantics, and all depend on dense, well-formed point clouds. PROSE instead operates at the object level, where a small set of high-confidence semantic correspondences absorbs the geometric noise that defeats descriptor-based pipelines.

\noindent\textbf{VLMs and foundation models for 3D scene understanding.} A growing literature builds 3D scene representations on top of pretrained vision-language and segmentation models~\cite{jung2026imagine3d,cho2024cat,kim2025seg4diff,shin2024towards,clip,yoon2025visual}. ConceptGraphs~\cite{gu2024conceptgraphs} and HOV-SG~\cite{werby2024hierarchical} construct open-vocabulary 3D scene graphs from multi-view RGB-D, while OpenScene~\cite{peng2023openscene} and OpenMask3D~\cite{takmaz2023openmask3d} lift CLIP features into dense 3D representations queryable in language. A separate thread prompts VLMs directly on 3D inputs: GPT4Scene~\cite{qi2025gpt4scene} grounds VLM reasoning in 3D scene tokens, while others~\cite{yang2023setofmark,gom2026aaai} promote spatial reasoning by overlaying marks on multimodal prompts. None of these lines, to our knowledge, targets the cross-scan correspondence task: scene-graph builders represent one scan, and VLM-prompting work reasons within one scene. PROSE composes both threads, using a single pretrained VLM to both build each scan's graph and match instances across them.

%% file: corl26/sections/3_method.tex
\section{Method}
\label{sec:method}
\vspace{-1mm}

\input{corl26/figures/2_architecture}

\noindent\textbf{Overview and problem setup.}
Given two egocentric RGB sequences $\mathcal{V}^{\mathrm{ref}}$ and $\mathcal{V}^{\mathrm{src}}$ of the same indoor scene captured at different times, we seek a rigid transform $T\in SE(3)$ aligning source to reference. \method requires no depth sensor and no externally provided SLAM trajectory as input: camera geometry is estimated from RGB by a geometric foundation backbone (Sec.~\ref{sec:method-representation}). The sequences may overlap only partially and contain moved or removed objects. As a byproduct of recovering $T$, we also output an instance-level open-vocabulary scene graph per scan, $\mathcal{G}^{\mathrm{ref}}$ and $\mathcal{G}^{\mathrm{src}}$, and the cross-scan instance correspondences $\mathcal{C}\subset V(\mathcal{G}^{\mathrm{ref}})\times V(\mathcal{G}^{\mathrm{src}})$ that drove the alignment. Figure~\ref{fig:main} summarizes \method: scene parsing (Sec.~\ref{sec:method-representation}), height-binned cross-scan correspondence and scene-graph matching (Sec.~\ref{sec:method-correspondence}), and pose hypothesis and voting (Sec.~\ref{sec:method-registration}).

\subsection{Scene representation: from RGB to per-instance 3D}
\label{sec:method-representation}

Scene parsing operates on each RGB sequence and produces a scene graph $\mathcal{G}=(V, E)$ whose nodes $V$ are 3D object instances and whose edges $E$ connect instances with nearby centroids.

\noindent\textbf{Per-frame geometry.} A geometric foundation model~\cite{wang2026vggt} takes the RGB sequence $\mathcal{V}$ and returns per-frame depth, intrinsics, and camera pose. We unproject and concatenate a subscan's frames into a single per-subscan point cloud with a 2D-to-3D index mapping each pixel to its 3D point, which later stages use to translate between per-frame masks and the unified point cloud.

\input{corl26/tables/1_prompts}

\noindent\textbf{Object listing.} A pretrained VLM~\cite{qwen3.6-27b} is prompted with a subsampled set of frames and returns a flat list of object names. We use a VLM rather than a closed-vocabulary detector~\cite{zhou2022detecting,liu2024grounding} because the relevant object set varies by scene, and because the prompt is designed for \emph{selection} rather than exhaustive detection: it elicits the objects most useful for cross-time alignment, not every visible object, since many add noise without aiding registration. The base prompt asks for at most 20 concrete, distinct items. On top of it, we layer five design choices targeting egocentric failure modes: (i) \emph{landmark salience}, prioritizing scene-anchoring furniture and fixtures; (ii) \emph{static-rigid focus}, skipping people, hands, and bare structural surfaces; (iii) a \emph{hallucination guard} forbidding speculative category variants; (iv) \emph{occlusion recovery}, retaining briefly or partially seen objects; and (v) a \emph{scene prior} that infers the room type and then its typical functional objects. The full prompt for each choice is given in Tab.~\ref{tab:prompt_variants}, and we merge them all for the final prompt.

\noindent\textbf{Instance segmentation.} Since the correspondence stage matches objects by instance identity rather than pixel-wise~\cite{cho2021cats,cho2022cats++,hong2021deep,hong2024unifying2,an2025cross} or point-wise~\cite{lee2026tora,lee2024pmtr, lee2025combinative,yue2025litept,hong2024pf3plat,han2025d,an2025c3g}, this step must give each physical object one persistent label across the sequence, not per-frame detections. We use SAM~3~\cite{carion2025sam}, which performs open-vocabulary concept segmentation: from a short noun phrase it aims to detect, segment, and track every instance of that concept across the video, yielding a temporally consistent mask $M^{(j)}_t$ for object $j$ and frame $t$. Unlike earlier segmenters~\cite{ravi2025sam} that accept only spatial prompts and return a single object, it consumes our VLM object names directly, avoiding the separate detector and tracker~\cite{cheng2022xmem,yang2022decoupling} that break under egocentric viewpoint change and occlusion. Identities can still split when an object leaves and re-enters the view; the fusion step below merges these back.

\noindent\textbf{Per-instance fusion.} The per-frame SAM~3 masks are not yet consistent across the sequence: the same physical object can take different instance IDs in different frames. To unify them, we back-project every masked pixel into 3D through the 2D-to-3D index (when nested masks claim a pixel, the smallest-area mask wins) and voxelize the combined cloud, so each voxel gathers points from many frames and potentially several IDs. We relabel each voxel's points to the locally majority ID, resolving per-frame ID switching into one label per region; an instance whose points are overwhelmingly revoted into another is then absorbed into it. A final scene-wide pass merges near-duplicate instances left by a split, using a geometric overlap test on their point clouds. The result is a scene graph $\mathcal{G}$ whose nodes are temporally consistent 3D instances, each carrying a fused point set and a PCA-based oriented bounding box, with edges connecting instances with nearby centroids. We provide full details in the appendix.

\subsection{Cross-scan instance correspondence}
\label{sec:method-correspondence}
Given the two scene graphs $\mathcal{G}^{\mathrm{ref}}$ and $\mathcal{G}^{\mathrm{src}}$, this stage produces instance correspondences $\mathcal{C}\subset V(\mathcal{G}^{\mathrm{ref}})\times V(\mathcal{G}^{\mathrm{src}})$. We again use a VLM here because indoor scenes contain many visually near-identical instances (multiple chairs, lamps, monitors) that confuse feature-similarity matchers, but a VLM can disambiguate by reasoning over surrounding context. Naive prompting still fails: the query is too large, and the VLM hallucinates matches between similar objects at different heights. Our fix is \emph{height-based binning}: comparing only instances at similar heights removes trivial distractors and constitutes the largest source of improvement in our ablation (Tab.~\ref{tab:ablation_corr}). We measure each instance's height along the gravity axis and split them into $K{=}5$ equal-count quantile bins with $20\%$ overlap between adjacent bins to retain object matches on the boundary. Querying the VLM separately within each bin keeps queries small and rules out implausible matches such as a ceiling lamp against a floor rug; a final pass recovers correspondences that never shared a bin.

Within each bin, we show the VLM two panels, one per scan, each a grid of that bin's object crops labeled with numbered Set-of-Marks~\cite{yang2023setofmark} markers from a single \emph{shared-namespace} pool: reference crops take markers $1\dots N$, and source crops take $N{+}1\dots N{+}M$. Every marker is unique across both panels, so a pair of marker numbers is already a cross-scan instance pair and the VLM's output is directly a list of candidate matches. We re-check each candidate with two prompts asking whether the objects are \emph{the same} and whether they are \emph{different}, keeping it only if the answers agree; since ``same?'' carries a yes-bias, the negated check exposes inconsistent hallucinations, sharply cutting false positives at a small recall cost. The surviving pairs form $\mathcal{C}$; examples are visualized in the appendix.

\subsection{Pose hypothesis and voting}
\label{sec:method-registration}

Given the instance correspondences $\mathcal{C}=\{(A_i, A'_i)\}_{i=1}^N$, where $(A_i, A'_i)$ is the $i$-th matched reference/source instance pair, and the per-subscan point clouds, we estimate the rigid transform $T\in SE(3)$ aligning source to reference. We assume a single dominant rigid motion relates the two scans, so one transform should align the shared, unmoved structure while moved objects act as outliers to be removed for reliable scene reconstructions~\cite{han2025emergent}. Rather than fitting $T$ to all correspondences jointly, we generate one transform hypothesis per matched pair and keep the one that best aligns the two point clouds, making the estimate robust to a minority of mismatched pairs.

For each matched pair, we fit a candidate transform $T_i$ by running RANSAC on the pair's point-level correspondences for outlier rejection, extracted between the two instances' fused point sets with an off-the-shelf descriptor~\cite{FCGF2019,rusu2009fast,qin2022geometric} under mutual NN matching. We then score each $T_i$ on its own, without pooling correspondences across pairs, by its inlier ratio: the fraction of the source point cloud that, once transformed by $T_i$, lands within a fixed distance of the reference cloud. The highest-scoring $T_i$ becomes the final $T$. A transform from a correctly matched pair aligns two scans, whereas one from a mismatched pair aligns little, so this selection is robust to individual bad pairs.
When no matched pair yields a usable hypothesis (\eg, when $\mathcal{C}$ is empty), PROSE falls back to registering the full point clouds directly with the same descriptor.

%% file: corl26/figures/2_architecture.tex
\begin{figure}
    \centering
    \includegraphics[width=1\linewidth]{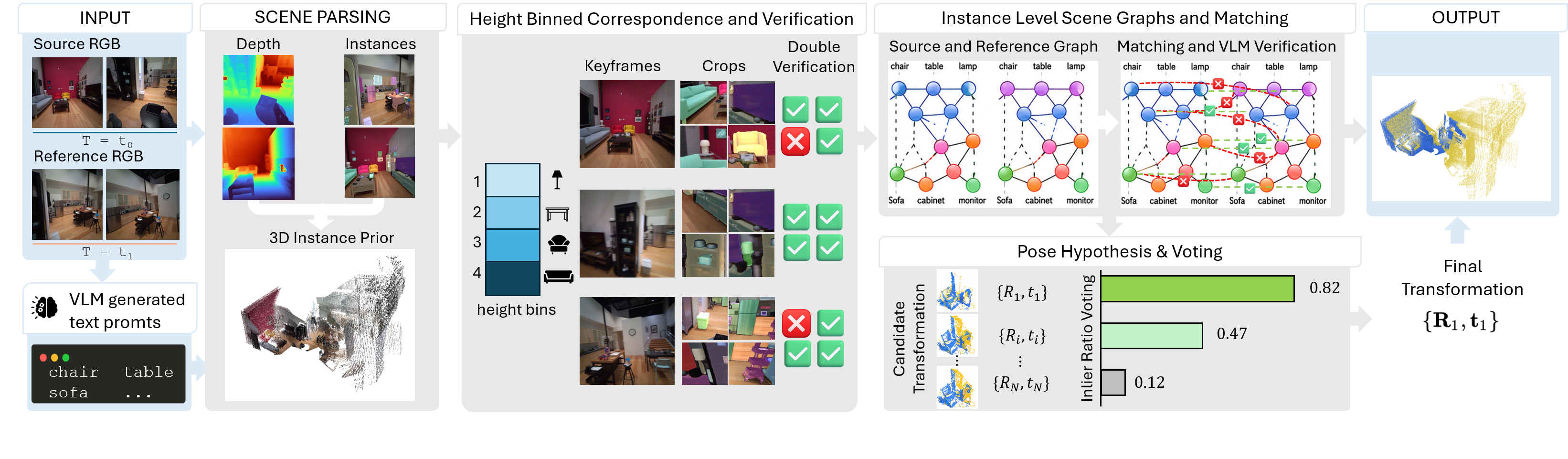}\vspace{-10pt}
   \caption{\textbf{Overall pipeline of \method.} From two egocentric RGB scans of the same scene at different times, \method parses each into a per-scan 3D scene graph, matches instances across scans, and estimates the rigid transform $T\in SE(3)$ by generating one candidate per matched pair and selecting the highest-inlier-ratio hypothesis. The entire pipeline is training-free.}
    \label{fig:main}
\end{figure}

%% file: corl26/tables/1_prompts.tex
\begin{table}[t]
\centering
\caption{\textbf{Object-listing prompt variants.} The Base object-listing prompt and the five design choices we ablate on top of it; each variant appends its instruction to the Base prompt.}
\label{tab:prompt_variants}
\small
\begin{tabularx}{\linewidth}{@{}lX@{}}
\toprule
\textbf{Variant} & \textbf{Prompt} \\
\midrule
Base & ``Identify all objects present in the video. For blurry frames, provide your best guess. Output at most 20 items as concrete, distinct objects you actually observe. Return the output as a Python list of strings.'' \\
\addlinespace
Landmark-Salience & ``Prioritize medium-to-large landmark-scale objects (furniture, appliances, built-in fixtures) that anchor the scene for relocalization, while still listing smaller distinctive items.'' \\
\addlinespace
Static-Rigid & ``Focus on static, rigid objects (furniture, appliances, fixtures, containers); skip non-rigid or dynamic entities (people, hands, pets) and bare structural surfaces (walls, floors, ceilings).'' \\
\addlinespace
Hallucination-Guard & ``Do not invent, speculate, or enumerate category variations (e.g., generic `toy X' / `small X' taxonomy expansions); never pad the list to reach the 20-item cap, and prefer fewer high-confidence items over many marginal guesses.'' \\
\addlinespace
Occlusion-Recovery & ``Include objects even when only briefly seen, partially visible, occluded, or in poor lighting---when in doubt, include them.'' \\
\addlinespace
Scene-Prior & ``First identify the room type (e.g., kitchen, living room, bedroom, office), then include functional objects typical of that room that you can directly see.'' \\
\bottomrule
\end{tabularx}
\end{table}

%% file: corl26/sections/4_experiments.tex
\section{Experiments}
\label{sec:experiments}
\vspace{-3pt}
\subsection{Implementation details}
\label{sec:exp-impl}
\vspace{-3pt}
We instantiate PROSE with VGGT-$\Omega$~\cite{wang2026vggt} as the geometric backbone, Qwen3.6-27B~\cite{qwen3.6-27b} for object listing and cross-scan instance matching, and SAM~3~\cite{carion2025sam} for text-prompted video segmentation; all are off-the-shelf checkpoints used without fine-tuning. For the final per-instance descriptor, we report results with FCGF~\cite{FCGF2019}, FPFH~\cite{rusu2009fast}, and GeoTransformer~\cite{qin2022geometric}, all pretrained and used as released. In per-instance fusion, we voxelize at $5$\,cm and resolve instance IDs by per-voxel majority vote, absorbing an instance into another when at least $50\%$ of its points are reallocated. For correspondence, we use $K{=}5$ quantile height bands with $20\%$ overlap. Registration uses RANSAC with 10K iterations and an inlier threshold of $3$\,cm. All experiments run on a single H200 GPU.

\input{corl26/tables/2_registration}
\subsection{Experimental Setting}
\label{sec:exp-setup}
\vspace{-3pt}
\noindent\textbf{Tasks and datasets.} We evaluate PROSE on three tasks: \emph{registration} recovers the $SE(3)$ transform aligning two subscans' point clouds; \emph{instance correspondence} forms ground-truth pairs by thresholding IoU under mutual nearest neighbors on mask-projected instances; and \emph{object listing} matches the predicted object list against a ground-truth list via Hungarian matching on CLIP~\citep{clip} cosine similarity. We use Aria Digital Twin (ADT)~\cite{pan2023aria} and Aria Everyday Activities (AEA)~\cite{lv2024aria}, two egocentric benchmarks of diverse indoor scenes. From 184 ADT sequences we derive 6{,}657 subscan pairs, organized into Single-ARIA (one wearer) and Multi-ARIA (multiple wearers) regimes, each further split by capture activity. ADT provides ground-truth clouds, poses, and object lists; AEA provides only poses and semi-dense clouds, so we evaluate AEA with VGGT-$\Omega$-predicted clouds. A simulated path-planning experiment (Sec.~\ref{sec:exp-pathplanning}) demonstrates downstream applicability.

\noindent\textbf{Evaluation metrics.} For registration we report Registration Recall (RR), the fraction of pairs with RRE $<5^\circ$ and RTE $<0.2$\,m; Relative Rotation/Translation Error (RRE/RTE); and Valid Ratio (VR), the fraction of pairs yielding a non-degenerate transform. For correspondence we report macro-averaged node precision, recall (NP, NR), F1, and coverage ratio (CR), defined as the fraction of pairs with $\geq 1$ correct matches. Because micro averaging over-rewards dense scenes, we use macro averaging and emphasize precision alongside coverage: precision drives registration quality, while coverage guarantees at least one strong match for the voting scheme (Sec.~\ref{sec:method-registration}) to select. For object listing we report precision, recall, and F1, plus anchor recall, all at a CLIP-similarity threshold of $0.5$ (denoted @50); anchor recall is computed over the ground-truth objects shared across both subscans, which are salient landmarks and make the best correspondence targets, while the other three are over all GT objects.

\subsection{Experimental Results}
\label{sec:exp-results}
\noindent\textbf{Scene registration.} Tab.~\ref{tab:adt_aea_main} compares cross-scan registration on ADT and AEA against scene-level baselines~\cite{yang2020teaser,qin2022geometric,Seo_BUFFERX_arXiv_2025} and the learned scene-graph baseline SG-Reg~\cite{11024207}. On ADT, we evaluate on both ground-truth clouds and clouds reconstructed from RGB by VGGT-$\Omega$, the sensor-free setting PROSE targets. On ground-truth clouds, PROSE leads in rotation error, translation error, and recall across all three ARIA splits, while SG-Reg, trained on curated indoor RGB-D, transfers poorly to egocentric data and trails every scene-level method. On reconstructed clouds, the baselines degrade sharply while PROSE retains usable accuracy and widens its lead, as object-level semantic correspondence is more robust to noisy, partial geometry than dense descriptor matching. No single descriptor backend dominates (GeoTransformer leads on reconstructed clouds, FCGF on ground truth), indicating that the gains stem from the correspondence stage. Notably, GeoTransformer runs out of memory at the scene level but works as a \method\ backend, since per-instance registration keeps each problem small.
\input{corl26/tables/3_ablations}

\input{corl26/tables/4_correspondence}
\noindent\textbf{Scene-graph correspondence.} We next evaluate the cross-scan instance matching that drives registration as a task in its own right (Tab.~\ref{tab:correspondence}), against an open-vocabulary detection-and-embedding baseline (GDino+CLIP) \cite{liu2024grounding} and SG-Reg \cite{11024207}. PROSE produces substantially more precise and higher-F1 correspondences than both baselines on ADT and AEA. The gap is largest in precision: matching object instances by VLM reasoning over shared-namespace Set-of-Marks prompts avoids the false matches that embedding-similarity methods make among visually similar objects, which is exactly the regime that dominates indoor scenes with repeated furniture. SG-Reg, whose learned matcher assumes curated graphs, produces few correct matches under the egocentric distribution, consistent with its weak registration numbers above.

\noindent\textbf{Ablation Study.} We justify our key design choices in object listing and correspondence in Tab.~\ref{tab:ablation_discovery} and Tab.~\ref{tab:ablation_corr} with evaluation on ADT dataset. Table~\ref{tab:ablation_discovery} ablates various prompting techniques and reveals the corresponding precision/recall tradeoff. Table~\ref{tab:ablation_corr} compares our method against two established spatial prompting techniques as well as isolates the contribution of each component in our pipeline. The two main contributors are the double verification and height binning, which sharpen precision and reduce distractors respectively. Additionally, we benchmark our object listing against open-vocabulary object tagging baselines in the appendix.

\subsection{Analysis: What Drives PROSE and Where It Breaks}
\label{sec:analysis}
\begin{figure}[t]
\centering
\begin{minipage}[t]{.32\linewidth}
    \vspace{0pt}
    \centering
    \captionof{table}{\textbf{Inlier ratio and RANSAC iterations for 99\% success.}}
    \label{tab:desc_ir}
    \footnotesize
    \setlength{\tabcolsep}{3pt}
    \begin{tabular}{lcc}
        \toprule
        Matching scope & IR & Iters \\
        \midrule
        Scene-level & 0.039 & ${\sim}$78K \\
        \rowcolor{gray!18}
        Matched objects & \textbf{0.137} & \textbf{${\sim}$1.8K} \\
        \bottomrule
    \end{tabular}
\end{minipage}%
\hfill
\begin{minipage}[t]{.32\linewidth}
    \vspace{0pt}
    \centering
    \captionof{table}{\textbf{Descriptor-free pose estimation.}}
    \label{tab:desc_free}
    \footnotesize
    \setlength{\tabcolsep}{3pt}
    \begin{tabular}{lcc}
        \toprule
        Method & RR & VR \\
        \midrule
        Centroid (Kabsch) & 32.3 & 36.5 \\
        OBB alignment & \second{56.5} & \second{89.5} \\
        \textbf{Ours$_{\text{GeoTr}}$} & \best{82.7} & \best{99.4} \\
        \bottomrule
    \end{tabular}
\end{minipage}%
\hfill
\begin{minipage}[t]{.31\linewidth}
    \vspace{0pt}
    \centering
    \includegraphics[width=.92\linewidth]{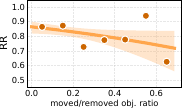}
    \vspace{-1mm}
    \captionof{figure}{\textbf{RR vs.\ change ratio.}}
    \label{fig:scene_change}
\end{minipage}
\vspace{-3mm}
\end{figure}

We probe where PROSE's accuracy comes from and how it degrades, examining the share
carried by the point descriptor, its robustness to scene change, and its sensitivity
to height binning. All analyses use the ADT Single-ARIA meal subset
(16 sequences, 561 pairs); absolute numbers on this split are higher than the
full-benchmark average in Tab.~\ref{tab:adt_aea_main}, so we focus on relative trends.

\textbf{Role of the point descriptor.} The final transform
(Sec.~\ref{sec:method-registration}) is estimated with the same descriptors that
underperform at the scene level; what differs is their operating point. With the
descriptor and inlier threshold fixed, restricting matching from the full scene to
a verified instance pair raises the inlier ratio (IR) of putative correspondences
from $0.039$ to $0.137$ (Tab.~\ref{tab:desc_ir}). Since the RANSAC iterations
required for $99\%$ success scale as $\log(0.01)/\log(1-\mathrm{IR}^3)$, the
requirement drops from ${\sim}78$K, infeasible under our 10K budget
(Sec.~\ref{sec:exp-impl}), to ${\sim}1.8$K. The gain thus stems from matching scope
rather than descriptor quality. Removing the descriptor altogether isolates its
share: aligning matched instances by their oriented bounding boxes alone still
reaches $56.5\%$ RR (Tab.~\ref{tab:desc_free}), and the descriptor adds $26.2$
points on top. Correspondence carries the registration; the descriptor refines it.

\textbf{Robustness to scene change.} We test whether PROSE fails catastrophically
as the scene changes between captures, evaluating RR as a function of the fraction
of objects moved or removed between the two captures (Fig.~\ref{fig:scene_change}).
RR declines gradually rather than collapsing, staying above $60\%$ even when nearly
two thirds of the objects have changed. Because inlier-ratio voting selects the
transform from a single matched pair, a few unmoved landmarks suffice to carry the
estimate.

\noindent
\begin{minipage}[t]{.61\linewidth}
\textbf{Sensitivity of height binning.} Height binning, the largest contributor in
our correspondence ablation (Tab.~\ref{tab:ablation_corr}), depends on two choices:
the bin count $K$ and the gravity axis, which we estimate as the minor principal
axis of each subscan cloud. As Tab.~\ref{tab:hb_k} shows, RR remains
stable over $K{=}3\text{--}7$, whereas removing binning ($K{=}1$) costs $13.7$ points. 

Tilting the estimated axis by $\Delta\theta$ degrades RR monotonically
(Tab.~\ref{tab:hb_tilt}), and at $15^\circ$ it falls back to roughly the no-binning
level ($68.8$ vs.\ $69.0$). Within the tested range, an erroneous gravity axis thus
forfeits the gain from binning rather than causing catastrophic failure.
\end{minipage}%
\hfill
\begin{minipage}[t]{.36\linewidth}
    \centering
    \vspace{-2mm}
    \captionof{table}{\textbf{Ablation on bin count $K$.}}
    \label{tab:hb_k}
    \footnotesize
    \setlength{\tabcolsep}{4pt}
    \begin{tabular}{lcccc}
        \toprule
        $K$ & 1 & 3 & \textbf{5 (Ours)} & 7 \\
        \midrule
        RR $\uparrow$ & 69.0 & \second{80.2} & \best{82.7} & 79.6 \\
        \bottomrule
    \end{tabular}

    \vspace{2mm}
    \captionof{table}{\textbf{Robustness to gravity-axis tilt $\Delta\theta$.}}
    \label{tab:hb_tilt}
    \footnotesize
    \setlength{\tabcolsep}{4pt}
    \begin{tabular}{lcccc}
        \toprule
        $\Delta\theta$ & \textbf{0$^\circ$ (Ours)} & $5^\circ$ & $10^\circ$ & $15^\circ$ \\
        \midrule
        RR $\uparrow$ & \best{82.7} & \second{76.8} & 69.4 & 68.8 \\
        \bottomrule
    \end{tabular}
\end{minipage}

\input{corl26/figures/3_path_planning}

\subsection{Downstream Application: Path Planning}
\label{sec:exp-pathplanning}
\vspace{-5pt}
\noindent\textbf{Setup.} We assess PROSE's scene graph by applying it to path planning. We capture a small room with an iPad at 30\,FPS, obtain depth and pose from VGGT-$\Omega$, and voxelize the back-projected cloud at 1\,cm for geometry and collision. Batched VLM object discoveries, merged into one list by a further VLM call, plus segmentation and voxel-revote deduplication, reduces the instance count from 258 (naive union) to 192. We connect each instance to its $k{=}5$ nearest neighbors within 1\,m and assign spatial relations (\emph{left-of}, \emph{near}, \emph{on top of}) from the recovered world frame. We sample node pairs at least $0.5$\,m apart, project colliding centroids to nearby free space, and plan the trajectory between each pair with Rapidly exploring Random Tree \citep{lavalle1998rapidly} (5\,s timeout) on a 2D slice $50$\,cm above the floor, emulating a quadruped such as Spot. Each query is scored as \emph{success}, \emph{no solution}, or \emph{outside room}. We display the environment with room constraint and a labeled sub-graph in Fig.~\ref{fig:ml_hall}.

\noindent\textbf{Results.} We report Success Rate (SR) by pair separation (near $0.5$--$0.75$\,m, far $0.75$--$1.25$\,m, cross-room $>1.25$\,m) and an in-room aggregate, with the pair, instance, and valid-instance counts each method yields. As displayed in Tab.~\ref{tab:path}, PROSE produces far more usable nodes than either baseline: FM-Fusion~\cite{fmfusion} recovers valid graphs for only a handful of rooms, so its perfect per-room rate reflects negligible coverage rather than planning quality, and ConceptGraphs~\cite{gu2024conceptgraphs} yields fewer valid targets across every band. The count columns characterize coverage more so than a preferred direction, as ground-truth graphs are unavailable; higher SR is unambiguously better.

%% file: corl26/tables/2_registration.tex
\newcolumntype{C}{>{\centering\arraybackslash}p{1.8cm}}

\newcommand{\best}[1]{\cellcolor{orange!20}\textbf{#1}}
\newcommand{\second}[1]{\cellcolor{orange!10}\underline{#1}}

\begin{table*}[t]
\centering
\caption{\textbf{Scene registration results on the ADT and AEA benchmarks.} The \colorbox{orange!20}{\textbf{best}} and \colorbox{orange!10}{\underline{second best}} results are highlighted within each setting. ($^{\dagger}$Max-clique solver times out on the very large putative correspondence set.)}
\label{tab:adt_aea_main}
\vspace{-5pt}
\setlength{\tabcolsep}{3pt}

\resizebox{0.98\textwidth}{!}{%
\begin{tabular}{l c ccc ccc cccc|cccc}
\toprule
& & \multicolumn{10}{c}{\textbf{ADT}} & \multicolumn{4}{|c}{\multirow{2}{*}{\textbf{AEA}}} \\
\cmidrule(lr){3-12}
& & \multicolumn{3}{c}{\textbf{Single-ARIA}} & \multicolumn{3}{c}{\textbf{Multi-ARIA}} & \multicolumn{4}{c}{\textbf{Total}} & \multicolumn{4}{|c}{} \\
\cmidrule(lr){3-5} \cmidrule(lr){6-8} \cmidrule(lr){9-12} \cmidrule(lr){13-16}
\textbf{Method} & \textbf{Type}
& \makecell{RR $\uparrow$ \\ (\%) } & \makecell{RRE $\downarrow$ \\ ($^{\circ}$)} & \makecell{RTE $\downarrow$ \\ (m) }
& \makecell{RR $\uparrow$ \\ (\%) } & \makecell{RRE $\downarrow$ \\ ($^{\circ}$)} & \makecell{RTE $\downarrow$ \\ (m) }
& \makecell{RR $\uparrow$ \\ (\%) } & \makecell{RRE $\downarrow$ \\ ($^{\circ}$)} & \makecell{RTE $\downarrow$ \\ (m) }
& \makecell{VR $\uparrow$ \\ (\%) }
& \makecell{RR $\uparrow$ \\ (\%) } & \makecell{RRE $\downarrow$ \\ ($^{\circ}$)} & \makecell{RTE $\downarrow$ \\ (m) }
& \makecell{VR $\uparrow$ \\ (\%) } \\
\midrule
\multicolumn{16}{c}{\cellcolor{gray!8}\textit{Using VGGT-$\Omega$ predicted point clouds}} \\
\midrule
TEASER++ $_\text{FPFH}$~\cite{yang2020teaser}  & \multirow{4}{*}{\makecell{Scene-level\\registration}} & 43.6 & 20.28 & 1.06 & 45.3 & 14.04 & 0.87 & 44.4 & 17.11 & 0.97 & \second{99.7} & 42.9 & 18.87 & 2.55 & \best{100} \\
TEASER++ $_\text{FCGF}$~\cite{yang2020teaser}  &  & \multicolumn{9}{c}{\cellcolor{gray!10} Fail to converge$^{\dagger}$} & 0 & \multicolumn{3}{|c}{\cellcolor{gray!10} Fail to converge$^{\dagger}$} & 0 \\
GeoTransformer~\cite{qin2022geometric}         &  & \multicolumn{9}{c}{\cellcolor{gray!25} Out of Memory (OOM)} & 0 & \multicolumn{3}{c}{\cellcolor{gray!25} Out of Memory (OOM)} & 0 \\
BUFFER-X \citep{Seo_BUFFERX_arXiv_2025}        &  & 35.3 & 20.28 & 1.17 & 38.9 & 13.73 & 0.90 & 37.1 & 16.95 & 1.03 & \best{100} & 34.5 & 20.35 & 2.44 & \best{100} \\
\cmidrule(lr){1-16}
SG-Reg \citep{11024207}                        & \multirow{4}{*}{\makecell{Scene-graph\\registration}} 
                                                & 20.7 & 29.57 & 3.27 
                                                & 17.1 & 29.61 & 3.27 
                                                & 19.1 & 29.58 & 3.27 & 76.3 
                                                & 19.4 & 27.06 & 3.52 & 52.6 \\
\textbf{Ours~$_\text{FPFH}$}                     &  & \second{51.2} & 18.33 & 0.80 & 48.8 & 13.23 & 0.61 & 49.9 & 15.71 & 0.70 & 95.5 & 47.9 & 17.91 & 1.84 & 87.3 \\
\textbf{Ours~$_\text{FCGF}$}                     &  & 49.3 & \best{14.64} & \second{0.69} & \second{50.6} & \second{11.87} & \second{0.57} & \second{50.0} & \second{13.21} & \second{0.63} & 95.5 & \second{58.5} & \best{9.04} & \best{0.99} & 88.3 \\
\textbf{Ours~$_\text{GeoTrans}$}                 &  & \best{55.1} & \second{14.71} & \best{0.66} & \best{57.3} & \best{10.49} & \best{0.51} & \best{56.2} & \best{12.55} & \best{0.59} & 96.0 & \best{65.5} & \second{12.71} & \second{1.39} & \second{89.6} \\
\midrule
\multicolumn{16}{c}{\cellcolor{gray!8}\textit{Using ground-truth (GT) point clouds}} \\
\midrule
TEASER++ $_\text{FPFH}$~\cite{yang2020teaser}  & \multirow{4}{*}{\makecell{Scene-level\\registration}} & 76.2 & 12.31 & 0.54 & 84.3 & 5.46 & 0.26 & 80.3 & 8.83 & 0.40 & \best{100} & \textendash & \textendash & \textendash & \textendash \\
TEASER++ $_\text{FCGF}$~\cite{yang2020teaser}  &  & 73.8 & 13.18 & 0.59 & 85.6 & 5.68 & 0.27 & 79.4 & 9.63 & 0.44 & \second{93.1} & \textendash & \textendash & \textendash & \textendash \\
GeoTransformer~\cite{qin2022geometric}          &  & \multicolumn{9}{c}{\cellcolor{gray!25} Out of Memory (OOM)} & 0 & \textendash & \textendash & \textendash & \textendash \\
BUFFER-X \citep{Seo_BUFFERX_arXiv_2025}         &  & 56.1 & 13.24 & 0.72 & 67.2 & 8.01 & 0.45 & 61.7 & 10.57 & 0.59 & \best{100} & \textendash & \textendash & \textendash & \textendash \\
\cmidrule(lr){1-16}
SG-Reg \citep{11024207}                         & \multirow{4}{*}{\makecell{Scene-graph\\registration}} 
                                                & 47.8 & 20.74 & 2.24 
                                                & 45.8 & 18.53 & 1.97 
                                                & 46.9 & 19.76 & 2.12 
                                                & 77.0 & \textendash & \textendash & \textendash & \textendash \\
\textbf{Ours~$_\text{FPFH}$}                     &  & \second{80.1} & \second{9.60} & \second{0.38} & \second{84.7} & \second{5.40} & \second{0.22} & \second{82.5} & \second{7.44} & \second{0.30} & \best{100} & \textendash & \textendash & \textendash & \textendash \\
\textbf{Ours~$_\text{FCGF}$}                     &  & \best{87.0} & \best{5.38} & \best{0.22} & \best{91.2} & \best{3.01} & \best{0.13} & \best{89.2} & \best{4.16} & \best{0.17} & \best{100} & \textendash & \textendash & \textendash & \textendash \\
\textbf{Ours~$_\text{GeoTrans}$}                 &  & 76.2 & 10.84 & 0.44 & 81.2 & 6.96 & 0.29 & 78.8 & 8.85 & 0.36 & \best{100} & \textendash & \textendash & \textendash & \textendash \\
\bottomrule
\end{tabular}%
\vspace{-10pt}
}
\end{table*}
\vspace{-5pt}

%% file: corl26/tables/3_ablations.tex
\begin{table}[!t]
\centering
\caption{\textbf{Ablation studies on \method}. We isolate the contribution of each stage's design choices.}

\label{tab:ablations}
\begin{subtable}[t]{0.73\linewidth}
\centering
\caption{Semantic object listing.}
\vspace{-5pt}
\label{tab:ablation_discovery}
\resizebox{\linewidth}{!}{%
\begin{tabular}{l ccccc cccc}
\toprule
\textbf{Ref} &
\makecell{Landmark-\\Salience} &
\makecell{Static-\\Rigid} &
\makecell{Hallucination-\\Guard} &
\makecell{Occlusion-\\Recovery} &
\makecell{Scene-\\Prior} &
\makecell{AR@50} & \makecell{P@50} & \makecell{R@50} & \makecell{F1@50} \\
\midrule
Base  & \textcolor{red}{\ding{55}} & \textcolor{red}{\ding{55}} & \textcolor{red}{\ding{55}} & \textcolor{red}{\ding{55}} & \textcolor{red}{\ding{55}} & 71.2          & \best{97.2}   & 65.2          & 72.5 \\
(a)   & \textcolor{green}{\ding{51}} & \textcolor{red}{\ding{55}} & \textcolor{red}{\ding{55}} & \textcolor{red}{\ding{55}} & \textcolor{red}{\ding{55}} & 78.1          & \second{95.8} & 73.5          & 77.3 \\
(b)   & \textcolor{green}{\ding{51}} & \textcolor{green}{\ding{51}} & \textcolor{red}{\ding{55}} & \textcolor{red}{\ding{55}} & \textcolor{red}{\ding{55}} & 87.6          & 94.1          & 82.5          & 85.3 \\
(c)   & \textcolor{green}{\ding{51}} & \textcolor{green}{\ding{51}} & \textcolor{green}{\ding{51}} & \textcolor{red}{\ding{55}} & \textcolor{red}{\ding{55}} & 90.3          & 94.8          & 85.0          & 88.3 \\
(d)   & \textcolor{green}{\ding{51}} & \textcolor{green}{\ding{51}} & \textcolor{green}{\ding{51}} & \textcolor{green}{\ding{51}} & \textcolor{red}{\ding{55}} & \second{91.5} & 95.5          & \second{85.9} & \second{89.0} \\
\textbf{Ours} & \textcolor{green}{\ding{51}} & \textcolor{green}{\ding{51}} & \textcolor{green}{\ding{51}} & \textcolor{green}{\ding{51}} & \textcolor{green}{\ding{51}} & \best{92.4} & \second{95.8} & \best{87.0} & \best{90.0} \\
\bottomrule
\end{tabular}%
}
\end{subtable}
\begin{subtable}[t]{0.212\linewidth}
\centering
\caption{Instance matching.}
\vspace{-5pt}
\label{tab:ablation_corr}
\resizebox{\linewidth}{!}{%
\begin{tabular}{l cc}
\toprule
\textbf{Method} & F1$\uparrow$ & CR$\uparrow$ \\
\midrule
GPT4Scene~\citep{qi2025gpt4scene} & 0.24          & \second{0.94} \\
GoM~\citep{gom2026aaai}           & \second{0.39} & \best{1.0}    \\
\midrule
\textbf{Ours}            & \best{0.45} & \best{1.0} \\
\quad-- cross-bin        & 0.44 & 1.0  \\
\quad-- 2$\times$ verif. & 0.36 & 1.0  \\
\quad-- cropping         & 0.34 & 0.83 \\
\quad-- height bins      & 0.29 & 0.58 \\
\bottomrule
\end{tabular}%
}
\end{subtable}
\hfill

\vspace{-10pt}
\end{table}

%% file: corl26/tables/4_correspondence.tex
\begin{wraptable}{r}{0.55\textwidth} 
    \vspace{-\intextsep} 
    \centering
    \caption{\textbf{Scene graph correspondence results on ADT and AEA.}}
    \label{tab:correspondence}
    \setlength{\tabcolsep}{3pt} 

    \resizebox{\linewidth}{!}{%
    \begin{tabular}{l cccc cccc}
    \toprule
    & \multicolumn{4}{c}{\textbf{ADT}} & \multicolumn{4}{c}{\textbf{AEA}}\\
    \cmidrule(lr){2-5} \cmidrule(lr){6-9}
    \textbf{Methods}
    & \makecell{NP $\uparrow$ \\ (\%)} & \makecell{NR $\uparrow$ \\ (\%)} & \makecell{F1 $\uparrow$ \\ (\%)} & \makecell{CR $\uparrow$ \\ (\%)}
    & \makecell{NP $\uparrow$ \\ (\%)} & \makecell{NR $\uparrow$ \\ (\%)} & \makecell{F1 $\uparrow$ \\ (\%)} & \makecell{CR $\uparrow$ \\ (\%)}\\ 
    \midrule

    GDino+CLIP \citep{liu2024grounding}      
                    & 14.2 & 23.6 & 17.2 & \second{89.7}
                    & \second{14.6} & \second{27.8} & \second{18.6} & \second{82.7} \\
    SG-Reg \citep{11024207}           
                    & \second{20.8} & \second{35.2} & \second{25.0} & 76.9
                    & 11.0 & 24.8 & 14.5 & 52.6 \\
    \textbf{Ours}   & \best{70.4} & \best{49.6} & \best{56.9} & \best{98.6} 
                    & \best{73.4} & \best{50.6} & \best{57.0} & \best{93.2}\\

    \bottomrule
    \end{tabular}
    }
    \vspace{-\intextsep} 
\end{wraptable}

%% file: corl26/figures/3_path_planning.tex
\begin{figure}[t]
    \centering
    \begin{subfigure}[c]{0.48\textwidth}
        \centering
        \refstepcounter{table}\label{tab:path}%
        \begin{minipage}{\linewidth}\small
        \textbf{Table~\thetable:}~\textbf{Path planning results.} We compare pairwise trajectory success rate and count statistics. *FM-Fusion has insufficient valid pairs $(n{=}2)$ to be considered competitive.
        \end{minipage}\par
        \vspace{4pt}
        \setlength{\tabcolsep}{2.5pt}
        \resizebox{\linewidth}{!}{%
        \begin{tabular}{l cccc ccc}
        \toprule
        & \multicolumn{4}{c}{\textbf{Success Rate} (\%) $\uparrow$} & \multicolumn{3}{c}{\textbf{Count}} \\
        \cmidrule(lr){2-5} \cmidrule(lr){6-8}
        \textbf{Method} & \makecell{SR$_\text{room}$} & \makecell{SR$_\text{near}$} & \makecell{SR$_\text{far}$} & \makecell{SR$_\text{cross}$}
        & \makecell{N$_\text{inst}$} & \makecell{N$_\text{pairs}$} & \makecell{N$_\text{room}$}\\
        \midrule
        FM-Fusion \citep{fmfusion}
                & 100* & 100* & 100* & -
                & 5 & 9 & 2\\
        ConceptGraphs \citep{gu2024conceptgraphs}
                & \second{46.2} & \second{47.1} & \second{46.8} & \second{38.1}
                & 33 & 379 & 234 \\
        \midrule
        \textbf{Ours}
                & \best{49.6} & \best{62.1} & \best{47.1} & \best{40.4}
                & 192 & 500 & 244 \\
        \bottomrule
        \end{tabular}}
        
        \vspace{6pt}
        \includegraphics[width=\linewidth]{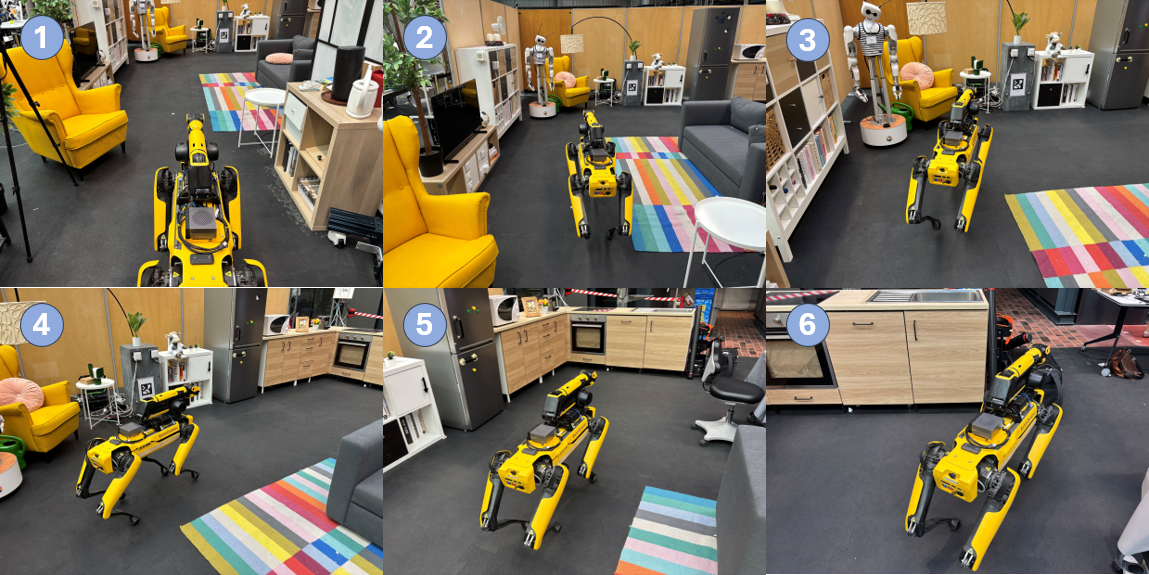}
        \caption{\textbf{Robotic demo.} Real-world assessment with robots can be found in the appendix.}
        \vspace{-3pt}
        \label{fig:robotic_demo}
    \end{subfigure}
    \hfill
    \begin{subfigure}[c]{0.48\textwidth}
        \centering
        \includegraphics[width=\linewidth]{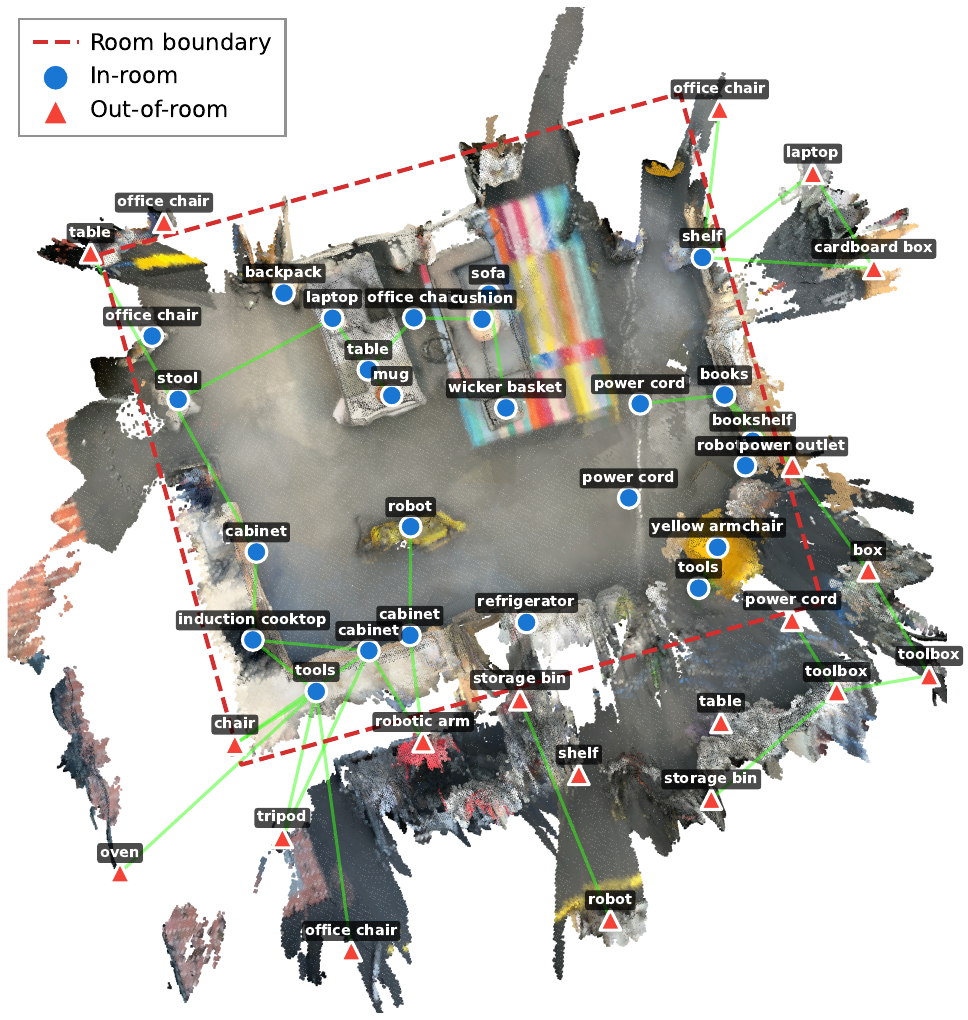}
        \caption{\textbf{Birds-eye-view of the open-plan room.}}
        \label{fig:right_bev}
    \end{subfigure}
    \vspace{2mm}
    \caption{\textbf{Visualization of path planning.} We show photographs of the experiment (a) alongside the reconstructed environment (b). We exclude objects outside the \textbf{\textcolor{red}{red box}} from the SR comparison.}
    \vspace{-4mm}
    \label{fig:ml_hall}
\end{figure}

%% file: corl26/sections/5_conclusion.tex
\section{Conclusion and Limitations}
\label{sec:conclusion}
We presented PROSE, a training-free pipeline for cross-time egocentric scene registration that uses a single pretrained VLM to build per-scan 3D scene graphs and match instances across captures. Height-compatible comparison and paired same/different verification make matching reliable enough for registration, with the final transform chosen by inlier-ratio voting. On Aria Digital Twin and Aria Everyday Activities, PROSE outperforms geometric and learned scene-graph baselines on ground-truth and RGB-reconstructed clouds, and its open-vocabulary scene graph supports simulated path planning. Limitations include VLM sensitivity to prompt phrasing, degraded performance under rapid motion or in sparse scenes, potential error accumulation across stages, and inference costs that preclude real-time use. Since each stage can be upgraded independently, PROSE should improve in accuracy and speed as foundation models advance.

\noindent\textbf{Acknowledgements.}
This work was supported by the Swiss AI Initiative through a grant from the Swiss National Supercomputing Centre (CSCS) under project ID ab036 on the Alps infrastructure; the Swiss National Science Foundation (Advanced Grant 216260: "Beyond Frozen Worlds: Capturing Functional 3D Digital Twins from the Real World"); the European Union’s Horizon Europe research and innovation programme under grant agreement No. 101214398 (ELLIOT); the Horizon Europe programme under grant agreement No. 101298421 (GRAIL); and the ETH AI Center through an ETH AI Center postdoc fellowship and an ETH Zurich Career Seed Award awarded to Sunghwan Hong.

%% file: corl26/sections/suppl.tex
\appendix
\renewcommand{\thefigure}{A\arabic{figure}}
\renewcommand{\thetable}{A\arabic{table}}
\setcounter{figure}{0}
\setcounter{table}{0}

\input{corl26/suppl/0_intro}
\input{corl26/suppl/1_real_robot}
\input{corl26/suppl/2_method_details}
\input{corl26/suppl/3_exp_details}
\input{corl26/suppl/4_additional_quan}
\input{corl26/suppl/5_additional_qual}

%% file: corl26/suppl/0_intro.tex
\newpage
\begin{center}
    \Large \textbf{Appendix}
\end{center}
\vspace{1em}

\noindent
In this appendix, we present additional information and analyses not included in the main paper. The contents are organized as follows:

\begin{tcolorbox}[
    enhanced, breakable,
    colback=gray!3, colframe=gray!40,
    boxrule=0.4pt, arc=2pt,
    left=10pt, right=10pt, top=8pt, bottom=8pt,
    title=\textbf{Contents},
    fonttitle=\normalsize, coltitle=black,
    colbacktitle=gray!12, attach boxed title to top left={yshift=-2pt, xshift=8pt},
    boxed title style={boxrule=0pt, arc=1pt}
]
\setlength{\parskip}{3pt}
\setlength{\parindent}{0pt}

\textbf{Section~\ref{sec:supp-real-robot}:~~Real-Robot Deployment and Path Planning.}~ Scene-graph setup, baseline comparisons, and the Spot navigation experiment.

\textbf{Section~\ref{sec:suppl-method-details}:~~Additional Method and Implementation Details.}~ Per-frame geometry, per-instance fusion, height-binned correspondence, and pairwise same/different verification.

\textbf{Section~\ref{sec:suppl-setup-details}:~~Experimental Setup Details.}~ Dataset construction and subscan pairing, task and metric definitions, baseline protocols, and hyperparameters.

\textbf{Section~\ref{sec:suppl-addtional-quan}:~~Additional Quantitative Results.}~ VLM backbone ablation, object-listing ablation, segmentation and fusion analysis, and end-to-end inference time.

\textbf{Section~\ref{sec:suppl-addtional-qual}:~~Additional Qualitative Results.}~ Registration and correspondence visualizations.
\end{tcolorbox}

%% file: corl26/suppl/1_real_robot.tex
\section{Real-robot Deployment and Path Planning}
\label{sec:supp-real-robot}

\subsection{Extended Path Planning Details}
\label{sec:supp-pathplanning}

\subsubsection{Setup}
\noindent\textbf{Scene graph construction.}
PROSE's scene graph is built from the full VGGT-$\Omega$~\cite{wang2026vggt} reconstruction (depth maps and camera poses) of 267 iPad frames captured at 30 FPS in a robotics laboratory. Batched VLM object listing queries discover 38 unique categories. SAM~3~\cite{carion2025sam} segments 258 instances and the subsequent voxel voting with deduplication yields the final 192 nodes. Edges are formed by connecting each node to its $k{=}5$ nearest neighbors within radius $r{=}1$\,m, producing 651 edges. The open-ended vocabulary includes domain-specific items such as \emph{robotic arm}, \emph{induction cooktop}, and \emph{wicker basket}. An optional VLM relabeling step, which has no effect on the path planning result, corrects SAM~3 errors from its top-4 frame context and the existing object list.

\noindent\textbf{Depth scale verification.}
VGGT-$\Omega$ predicts depth at an arbitrary scale with no explicit metric supervision. We verify that its predicted scale is approximately metric by comparing the bounding diagonal of the VGGT-$\Omega$ back-projected cloud (7.2M points, 5\,mm voxel stride) against two independent references: a metric-scale cloud from DA3~\cite{lin2025depth}, a monocular depth model trained to predict metric depth directly, and the ground-truth mesh from the iOS 3D Scanner App. The VGGT-$\Omega$ diagonal (12.81\,m) matches DA3 (12.81\,m) almost exactly and exceeds the GT mesh (12.27\,m) by 4\%, the latter gap arising because the iOS LiDAR is depth-capped and truncates geometry beyond the immediate room, whereas VGGT-$\Omega$ recovers it. Pairwise centroid distances between the 192 instances common to VGGT-$\Omega$ and DA3 correlate at Pearson $r{=}0.999$, and $k{=}5$ neighborhood overlap is 93\% (131/192 perfect). We therefore use VGGT-$\Omega$'s predicted depth and poses directly, without post-hoc scale correction.

\noindent\textbf{TSDF Obstacle Map and Coordinate Frame.}
The VGGT-$\Omega$ depth maps are integrated via Open3D~\cite{zhou2018open3d} TSDF at 5\,mm resolution, producing a colored mesh with 660K vertices and 1.2M triangles. For collision checking, the mesh is voxelized at 2\,cm ($\approx$150K obstacle voxels). Because the predicted frame is not gravity-aligned, we compute the PCA frame of the mesh vertices and all obstacle voxels and node centroids are transformed to this PCA frame before planning. The 2D obstacle map is constructed by extracting voxels in a $z$-band around the robot navigation height ($z{=}0.0 \pm 0.05$ in PCA coordinates, corresponding to ${\approx}50$\,cm above the physical floor). This yields ${\sim}7{,}500$ obstacle voxels for collision queries via a KD-tree.

\noindent\textbf{Planning Protocol.}
We wrap OMPL's~\cite{sucan2012open} bidirectional RRT-Connect~\cite{lavalle1998rapidly} with KD-tree–based collision checking. The planner operates in obstacle-avoidance mode: a state is valid if and only if its distance to the nearest obstacle voxel exceeds the inflation radius $r_{\mathrm{inv}}{=}0.06$\,m. A user-defined room polygon in PCA $XY$ (four corners hand-picked from the BEV) is enforced inside the state-validity checker such that paths that exit the polygon are invalid. After finding a solution, OMPL's \texttt{simplifySolution()} shortens the path. The timeout is 5s per pair.

Since node centroids typically lie inside their parent object's solid geometry, we snap each start/goal to the nearest collision-free point via a radial search (16 angles per ring, 2\,cm step, up to 50\,cm) constrained to the room polygon. The polygon is expanded by the snap margin (10\,cm) for the initial in-room gate, so nodes near walls are not prematurely excluded. Each pair passes through a pre-flight failure cascade before the expensive RRT solve:
\begin{enumerate}[nosep]
\item \emph{Outside room}: centroid $XY$ falls outside the expanded room polygon $\rightarrow$ skipped.
\item \emph{Unreachable}: centroid is in-room but embedded in solid geometry; the snap-to-free-space search fails $\rightarrow$ skipped.
\item \emph{No solution}: both endpoints are valid free-space positions, but RRT times out or the path exits the room boundary.
\item \emph{Success}: collision-free path found entirely within the room.
\end{enumerate}

\subsubsection{Baselines}
All methods use the same protocol: enumerate all $\binom{N}{2}$ node pairs, discard pairs whose PCA $XY$ centroid distance is below 0.5\,m (trivially short). This yields 500 pairs for PROSE, 379 for ConceptGraphs and 9 for FM-Fusion~\cite{fmfusion}. All methods share the same VGGT-$\Omega$ TSDF obstacle mesh, room polygon, and planner parameters.

\noindent\textbf{ConceptGraphs.}
We run the three-stage ConceptGraphs~\cite{gu2024conceptgraphs} pipeline: (i)~RAM~\cite{zhang2024recognize} + GroundingDINO~\cite{liu2024grounding} + SAM (ViT-H)~\cite{kirillov2023segment} on 267 frames; (ii)~\texttt{cfslam\_pipeline\_batch} mapping with overlap-based spatial similarity (\texttt{sim\_threshold=1.0}, \texttt{dbscan\_eps=0.1}, \texttt{min\_detections=3}); (iii)~node extraction at $\geq50$ points, edges via $k$-NN ($k{=}5$, $r{=}1.0$\,m). This yields 33 nodes and 165 edges. ConceptGraphs uses the same VGGT-$\Omega$ depth and poses as PROSE, so node centroids and the TSDF obstacle mesh are in the same coordinate frame.

\noindent\textbf{FM-Fusion.}
SG-Reg's~\cite{11024207} relative underperformance on our benchmarks can be attributed to its scene graph generation backbone, FM-Fusion~\cite{fmfusion}, as highlighted in this experiment. 
FM-Fusion's Bayesian label fusion requires frame-to-frame depth consistency for 3D IoU–based instance tracking. With VGGT-$\Omega$ estimates, per-frame depth drift causes nearly all cross-frame associations to fail: only 5 instances survive (\emph{bookshelf}, \emph{ceiling}, \emph{couch}, \emph{floor}, \emph{fridge}) out of $250{+}$ single-frame fragments created. The label vocabulary is closed to ${\sim}20$ NYUv2/ScanNet classes: domain-specific objects are coerced to the nearest category (\emph{robotic arm} $\to$ \emph{robot}) or dropped entirely.

\subsubsection{Coverage Analysis}
FM-Fusion's~\cite{fmfusion} 2/2 in-room success rate reflects the fact that its 5 surviving nodes are large objects whose centroids happen to sit in navigable space but 7 of 9 pairs fall outside the room polygon, leaving negligible coverage for routing between arbitrary objects. ConceptGraphs~\cite{gu2024conceptgraphs} recovers 33 nodes, but its class-agnostic detection plus aggressive spatial merging biases survival toward large, high-confidence objects: the resulting centroids cluster around a few dominant structures (\emph{tables}, \emph{bookshelves}, \emph{cabinets}), leaving the room interior sparsely covered and domain-specific items (\emph{robotic arm}, \emph{induction cooktop}) absent. PROSE at 192 nodes provides both the densest spatial coverage and the richest semantic target set, yielding the broadest set of navigable routing targets among the three methods.

\subsection{Real-Robot Experiment}
We validate PROSE in a real-world object navigation scenario using a Boston Dynamics Spot robot equipped with an arm. The experiments are conducted in an indoor environment to evaluate PROSE's ability to support object-level navigation. For each navigation task, we use the PROSE scene graph constructed from the iPad scan, as described previously, and localize the robot within the same map using an image-based localization pipeline~\cite{sarlin2019coarse}. This provides the robot's current position and orientation in the scene graph.

Given a human query specifying a target object in the scene, the planning pipeline introduced earlier computes a path from the robot's current location to the nearest collision-free waypoint associated with the target object. The resulting path is collision-free by construction. Spot then tracks the interpolated waypoints along the planned path using the built-in velocity controller provided by the official Spot SDK. In addition, Spot's local collision-avoidance functionality further reduces the risk of collisions during execution. The experiments demonstrate that PROSE enables a mobile robot to successfully navigate to target objects and places in real indoor scenes. 

%% file: corl26/suppl/2_method_details.tex
\section{Additional Method/Implementation Details}
\label{sec:suppl-method-details}

This section expands the compressed method description of the main paper (Sec.~\ref{sec:method}) into the concrete detail needed to re-implement the stages whose implementation choices are non-obvious: per-frame geometry and per-instance fusion within per-scan scene representation (Sec.~\ref{sec:method-representation}), and cross-scan correspondence with its pairwise verification (Sec.~\ref{sec:method-correspondence}).

\subsection{Details on Per-Frame Geometry}
\label{sec:suppl-perframe}

Here, we detail how \method recovers the per-subscan point clouds that every later stage operates on, expanding the per-frame geometry step of Sec.~\ref{sec:method-representation}. 
\method assumes no depth sensor and no externally provided trajectory: all camera geometry is predicted from RGB.

\begin{wrapfigure}{r}{0.54\textwidth}
    \centering
    \vspace{-4mm}
    \includegraphics[width=\linewidth]{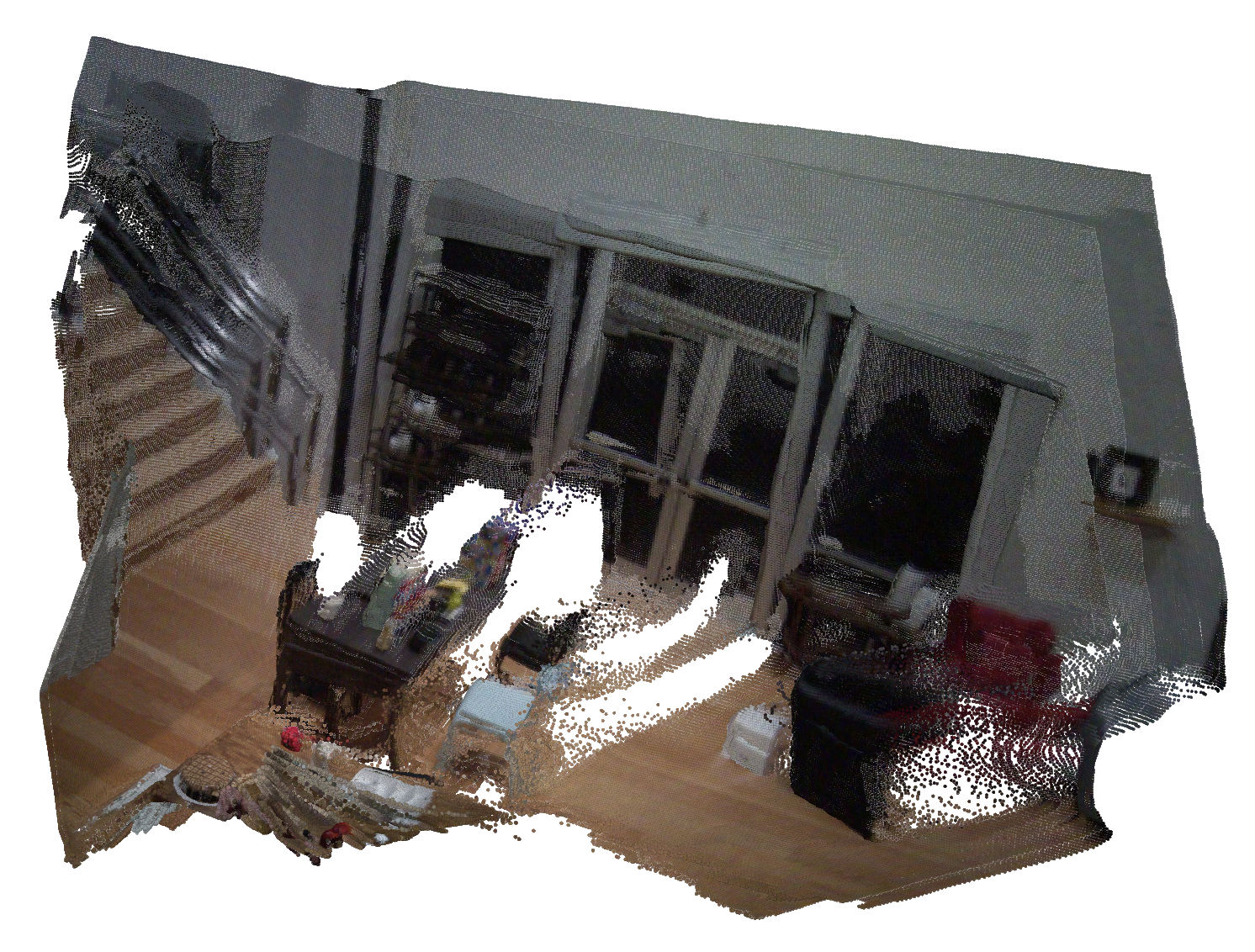}
    \caption{\textbf{Per-subscan point cloud example.} An example subscan point cloud predicted by VGGT-$\Omega$.}
    \vspace{-2mm}
    \label{fig:persubscan_pc}
\end{wrapfigure}

Each subscan is a short RGB sequence $\mathcal{V}$. We pass $\mathcal{V}$ through the geometric foundation backbone VGGT-$\Omega$~\cite{wang2026vggt}, used off the shelf without fine-tuning, which predicts a dense depth map, the camera intrinsics, and the camera pose for every frame. Writing $d_f$ for the predicted depth, $K_f$ for the predicted intrinsics, and $(u,v)$ for a pixel of frame $f$, the per-frame back-projection produces a point $\mathbf{X}^{\mathrm{cam}}_{f}(u,v)\in\mathbb{R}^3$ in that frame's camera frame:
\begin{equation}
\mathbf{X}^{\mathrm{cam}}_{f}(u,v) \;=\; d_f(u,v)\,K_f^{-1}\,[u,\,v,\,1]^{\top},
\label{eq:suppl-unproject}
\end{equation}
applied per pixel. We then concatenate the per-frame clouds of a subscan into a single per-subscan point cloud (Fig.~\ref{fig:persubscan_pc}). However, this plain concatenation keeps only the 3D coordinates and forgets which frame and pixel each point was projected from, so we retain a \emph{2D-to-3D index} that links every 3D point to its source pixel and, conversely, every pixel to the 3D point it produced. Later stages use this index in both directions: to carry per-frame SAM~3~\cite{carion2025sam} mask labels onto the unified cloud (Sec.~\ref{sec:suppl-fusion}), and to trace a 3D instance back to the frames and pixels in which it appears (Sec.~\ref{sec:suppl-binning}).

\subsection{Details on Per-Instance Fusion}
\label{sec:suppl-fusion}

\input{corl26/suppl/figures/2_fusion_example}

As shown in Fig.~\ref{fig:fusion_failure}, the segmentation stage (Sec.~\ref{sec:method-correspondence}) emits per-frame SAM~3~\cite{carion2025sam} masks, but their instance identities are not yet consistent across the sequence: the same physical object may receive different IDs in different frames, and a single object may be split when it leaves and re-enters view. Fusion turns these per-frame masks into the temporally consistent 3D instances of the scene graph $\mathcal{G}$ (Alg.~\ref{alg:suppl-fusion}).

\renewcommand{\algorithmicrequire}{\textbf{Input:}}
\renewcommand{\algorithmicensure}{\textbf{Output:}}
\begin{algorithm}[h]
\caption{Per-Instance Fusion: per-frame masks $\rightarrow$ scene graph $\mathcal{G}$}
\label{alg:suppl-fusion}
\begin{algorithmic}[1]
\REQUIRE per-frame masks $\{M_f\}$; per-subscan cloud with 2D-to-3D index (Sec.~\ref{sec:suppl-perframe})
\ENSURE scene graph $\mathcal{G}=(V,E)$
\FORALL{frame $f$ and masked pixel $p$}
    \STATE assign $p$ to the smallest-area mask covering it
    \STATE back-project $p$ to a 3D point via the 2D-to-3D index
\ENDFOR
\STATE pool the labeled points over all frames into one cloud $C$ (a region may hold several IDs)
\STATE voxelize $C$ at $5$\,cm
\FORALL{occupied voxel}
    \STATE relabel its points to the locally majority instance ID
\ENDFOR
\FORALL{instance $a$}
    \IF{at least $50\%$ of $a$'s points are revoted to a single instance $b$}
        \STATE merge $a$ into $b$
    \ENDIF
\ENDFOR
\FORALL{instance pair $(a,b)$}
    \STATE compute $\rho_{\mathrm{vox}}(a,b)$ and $d_{\mathrm{cham}}(a,b)$ (Eq.~\ref{eq:suppl-dedup-vox} and~\ref{eq:suppl-dedup-cham})
    \IF{$\rho_{\mathrm{vox}}(a,b) \ge 0.6$ and $d_{\mathrm{cham}}(a,b) \le 3$\,cm}
        \STATE mark $(a,b)$ for merging
    \ENDIF
\ENDFOR
\STATE merge each transitively-connected group of marked pairs into one node
\STATE $V \gets$ fused instances, each with a PCA oriented bounding box
\STATE $E \gets$ edges between instances with nearby centroids
\STATE \textbf{return} $\mathcal{G}=(V,E)$
\end{algorithmic}
\end{algorithm}

\textbf{Lifting masks to 3D.} For each frame we back-project every masked pixel into 3D through the 2D-to-3D index of Sec.~\ref{sec:suppl-perframe}. Within a frame, masks can be nested; when several masks claim the same pixel, the smallest-area mask wins, so the more specific object takes the point. Pooling the lifted points across all frames yields one combined cloud in which each surface region carries votes from many frames and, because identities are not yet unified, potentially several instance IDs.

\textbf{Per-voxel majority voting.} We voxelize the combined cloud at $5$\,cm. Each occupied voxel thus aggregates points contributed by multiple frames and IDs, and we relabel all points in a voxel to the locally majority instance ID. This collapses per-frame ID switching into a single coherent label per spatial region, the key step that stitches a physical object's scattered per-frame identities back together.

\textbf{Absorption.} Majority voting can leave a fragment ID whose points have largely been revoted into a neighbor. When at least $50\%$ of an instance's points are reallocated to a single other instance, we absorb the former into the latter, removing the redundant label. This cleans up the over-segmentation that remains after voxel voting.

\textbf{Scene-wide near-duplicate merge.} A final global pass targets duplicates that voxel voting cannot reach: two instances that describe the same object but never co-occupy voxels, typically the two halves of an object split when it left and re-entered the view. For every pair of instances $a,b$ we apply a two-part geometric overlap test on their $5$\,cm voxelizations $V_a, V_b$ and on the two one-directional mean nearest-neighbor distances $d_{a\to b}$ and $d_{b\to a}$ between their point clouds. The first gate is a \emph{voxel-containment} test on the smaller instance's share of shared voxels,
\begin{equation}
\rho_{\mathrm{vox}}(a,b) \;=\; \frac{|V_a\cap V_b|}{\min(|V_a|,|V_b|)} \;\ge\; \tau_{\mathrm{vox}},
\qquad \tau_{\mathrm{vox}} = 0.6,
\label{eq:suppl-dedup-vox}
\end{equation}
which catches a fragment contained in its full counterpart while ignoring how the two clouds differ in extent. The second is a \emph{Chamfer} gate that guards against two distinct objects whose bounding regions merely overlap,
\begin{equation}
d_{\mathrm{cham}}(a,b) \;=\; \min\!\bigl(d_{a\to b},\,d_{b\to a}\bigr) \;\le\; \tau_{\mathrm{cham}},
\qquad \tau_{\mathrm{cham}} = 3\,\mathrm{cm},
\label{eq:suppl-dedup-cham}
\end{equation}
requiring at least one cloud to lie on average within $3$\,cm of the other. A pair is merged only when \textbf{both} Eq.~(\ref{eq:suppl-dedup-vox}) and Eq.~(\ref{eq:suppl-dedup-cham}) hold; qualifying pairs are grouped transitively, so one physical object becomes one node.

\textbf{Output.} The result is the scene graph $\mathcal{G}=(V,E)$:
\begin{itemize}
  \item Nodes $V$: temporally consistent 3D instances; each node carries its fused point set.
  \item Edges $E$: each instance is connected with its $k{=}5$ nearest neighbors within 1\,m.
\end{itemize}

We visualize reconstructed scene graph examples in Sec.~\ref{sec:suppl-addtional-qual}.


\subsection{Details on Height-Binned Correspondence}
\label{sec:suppl-binning}

Given the two scene graphs $\mathcal{G}^{\mathrm{ref}}$ and $\mathcal{G}^{\mathrm{src}}$, the correspondence stage (Sec.~\ref{sec:method-correspondence}) matches object instances across scans with the VLM (\eg, Qwen3.6-27B~\cite{qwen3.6-27b}). A single query over all instances is both too large and prone to hallucinated matches between similar objects at different heights; \emph{height-based binning} is our fix and the largest single contributor in the correspondence ablation (Tab.~\ref{tab:ablations}). Here we provide the details on the binning and the prompt layout. 

\begin{wrapfigure}{r}{0.48\textwidth}
    \centering
    \vspace{-3mm}
    \includegraphics[width=\linewidth]{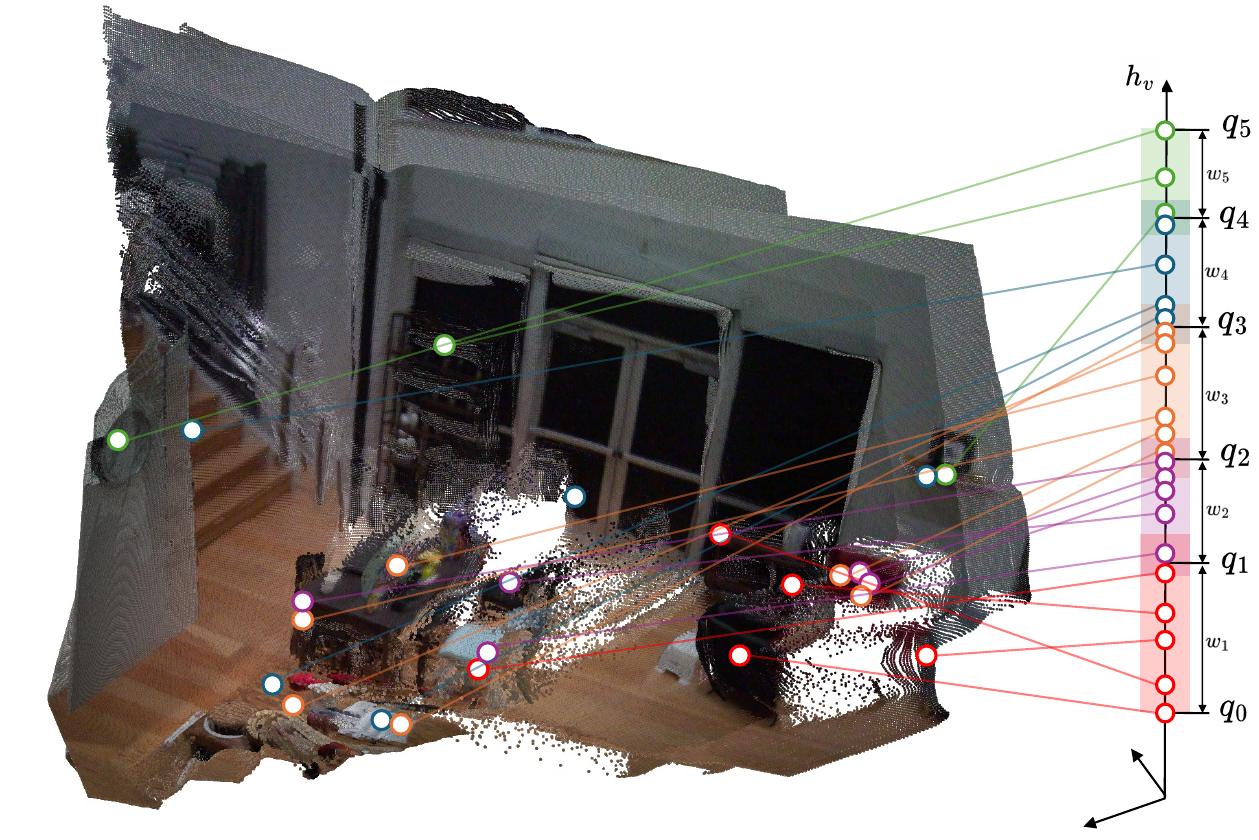}
    \caption{\textbf{Height binning on the gravity axis.} Each object instance is placed on the gravity ($h_v$) axis at its centroid height (colored markers, linked to the axis). The pooled heights are cut into $K{=}5$ \emph{equal-count quantile} bins, with edges $q_0{=}h_{\min},\dots,q_5{=}h_{\max}$ and native widths $w_1,\dots,w_5$ (Eq.~\ref{eq:suppl-bin-edges}). Each bin is then expanded across its edges by $\alpha{=}0.2$ of the smaller adjacent width into the overlapping bands $B_k$ (Eq.~\ref{eq:suppl-bin}), so an instance near a boundary still falls in a shared bin and stays matchable.}
    \vspace{20mm}
    \label{fig:bin}
\end{wrapfigure}

\textbf{Height bins.} We measure each instance's height as the coordinate of its centroid along the gravity axis, and represent it with $h_v$. We estimate this axis per subscan as the minor principal axis of the point cloud, with its sign resolved by the mean camera up-vector. We form $K{=}5$ \emph{equal-count quantile} bins over the heights pooled from both scans, then expand each bin across its edges so neighbors overlap (Fig.~\ref{fig:bin}). Let
\begin{multline}
q_k = \mathrm{Quantile}\!\left(\{h_v\}_{v\in V^{\mathrm{ref}}\cup V^{\mathrm{src}}},\; k/K\right),
\quad k=0,\dots,K,
w_k = q_k - q_{k-1},\quad k=1,\dots,K,
\label{eq:suppl-bin-edges}
\end{multline}
 
where $q_0,\dots,q_K$ are the $K{+}1$ \emph{quantile edges} and each $w_k=q_k-q_{k-1}$ is the width of the $k$-th \emph{native} bin $[q_{k-1},q_k]$, \ie, its width before any overlap is added. With overlap fraction $\alpha = 0.2$, the buffer applied at each edge is $\alpha$ times the \emph{smaller} of the two widths meeting at that edge (so a sparse, wide bin cannot engulf a narrow neighbor), and the outermost edges use the bin's own width and are clipped to the observed height range:
\begin{equation}
B_k =
\Bigl[\,
\max\!\bigl(h_{\min},\; q_{k-1} - \alpha\, e^{-}_k\bigr),
\;\;
\min\!\bigl(h_{\max},\; q_{k} + \alpha\, e^{+}_k\bigr)
\,\Bigr],
\qquad
\begin{aligned}
e^{-}_k &= \begin{cases} w_1 & k=1\\ \min(w_k, w_{k-1}) & k>1\end{cases}\\[2pt]
e^{+}_k &= \begin{cases} w_K & k=K\\ \min(w_k, w_{k+1}) & k<K,\end{cases}
\end{aligned}
\label{eq:suppl-bin}
\end{equation}

where $h_{\min}=\min_v h_v$ and $h_{\max}=\max_v h_v$. \emph{Before} overlap, each native quantile bin $[q_{k-1},q_k]$ holds roughly the same number of pooled instances regardless of how heights are distributed (exact equality is precluded by tied heights). An instance then joins every \emph{expanded} bin $B_k$ that contains its centroid height and may appear in more than one, so the queried bins deliberately differ in count, and this duplication keeps boundary objects from being lost: an object sitting near a bin edge, or one whose two scans place its centroid slightly differently, still falls inside a shared bin and remains matchable.

\input{corl26/suppl/figures/4_corr_input}

\input{corl26/suppl/tables/correspondence_prompt}

\textbf{Per-bin querying.} We query the VLM separately within each bin. Comparing only similar-height instances keeps each query small and removes implausible candidates outright, such as a ceiling lamp matched against a floor rug. Because the overlapping quantile bins do not by themselves cover every pair, a final cross-bin pass re-queries the instances that were left \emph{unmatched} by every per-bin query, whether because they never co-occurred in a shared bin or because the VLM simply missed them, grouping them into one additional catch-all query and recovering correspondences the per-bin queries could not see. 

\textbf{Shared-namespace Set-of-Marks.} Within a bin we present the VLM with three images (Fig.~\ref{fig:s4_input}) and engineer the prompt layout so that the VLM's output is directly a list of candidate cross-scan pairs:
\begin{itemize}
  \item A \emph{REF crop grid} and a \emph{SRC crop grid}: one zoomed crop per object in that bin, each crop centered on a single instance whose silhouette is outlined in green and whose numeric Set-of-Marks~\cite{yang2023setofmark} ID is stamped at the crop's top-left corner.
  \item A \emph{context frame}: one wide RGB frame per scan (REF on top, SRC on bottom) showing those same objects in their spatial layout, outlined and labelled with the same IDs, so the model can confirm a match from neighbours and arrangement rather than appearance alone.
  \item A single shared marker pool: reference crops take markers $1\dots N$, source crops take markers $N{+}1\dots N{+}M$; every marker number is unique across both scans.
  \item A marker pair returned by the VLM is therefore directly a cross-scan instance pair, with no separate alignment step; these candidates are filtered by the verification of Sec.~\ref{sec:suppl-verification}.
\end{itemize}

Tab.~\ref{tab:suppl-prompts-som} reproduces the system and user prompts for this per-bin query verbatim; the reported configuration uses the non-thinking-mode, crops-plus-context variant shown here.

\subsection{Pairwise Same/Different Verification}
\label{sec:suppl-verification}

\input{corl26/suppl/tables/verification_prompt}
\input{corl26/suppl/figures/5_corr_output}

Each marker pair from the binning stage (Sec.~\ref{sec:suppl-binning}) is only a candidate. This subsection details the pairwise verification that filters candidates into the final correspondence set $\mathcal{C}$; together with height binning it is one of the two largest contributors in the correspondence ablation (Tab.~\ref{tab:ablation_corr}).

\textbf{Verification gate.} For each candidate pair we render the two instances side by side (source crop on the left, reference crop on the right) into a single image and pose two complementary yes/no questions to the VLM:
\begin{itemize}
  \item \emph{Affirmative prompt}: asks whether the left and right crops are the same physical object; we keep the pair if the answer is ``same''.
  \item \emph{Negated prompt}: asks whether they are different objects; we keep the pair if the answer is ``not different''.
\end{itemize}
Let $\pi_{\mathrm{VLM}}$ denote Qwen3.6-27B viewed as a yes/no answerer over the rendered pair. The negated prompt is evaluated only on the pairs that survive the affirmative prompt, and a candidate is retained only when both answers agree, formalized as
\begin{equation}
(A_i, A'_i) \in \mathcal{C}
\;\iff\;
\pi_{\mathrm{VLM}}(\mathrm{same}\,?\,\mid A_i,A'_i) = \texttt{yes}
\;\land\;
\pi_{\mathrm{VLM}}(\mathrm{different}\,?\,\mid A_i,A'_i) = \texttt{no}.
\label{eq:suppl-verify}
\end{equation}

\textbf{Why two prompts.} The two prompts are deliberately redundant. A bare ``are these the same?'' query carries a yes-bias, so the VLM tends to confirm weak matches; the negated phrasing of the second prompt does not share that bias, and a hallucinated match that the affirmative prompt accepts is frequently rejected by the negated one. Requiring agreement therefore exposes inconsistent answers and sharply cuts false positives, at a small cost in recall. The pairs that survive both prompts constitute the cross-scan correspondence set $\mathcal{C}$ used for pose estimation (Sec.~\ref{sec:method-registration}); Fig.~\ref{fig:s4_output} shows example candidate pairs accepted into $\mathcal{C}$ and those rejected by this gate.

Table~\ref{tab:suppl-prompts-verify} reproduces the four prompts verbatim, with the affirmative pair implementing the \texttt{same?} query and the negated pair implementing the \texttt{different?} query of Eq.~(\ref{eq:suppl-verify}).

%% file: corl26/suppl/figures/2_fusion_example.tex
\begin{figure}[h]
    \centering
    \includegraphics[width=1\linewidth]{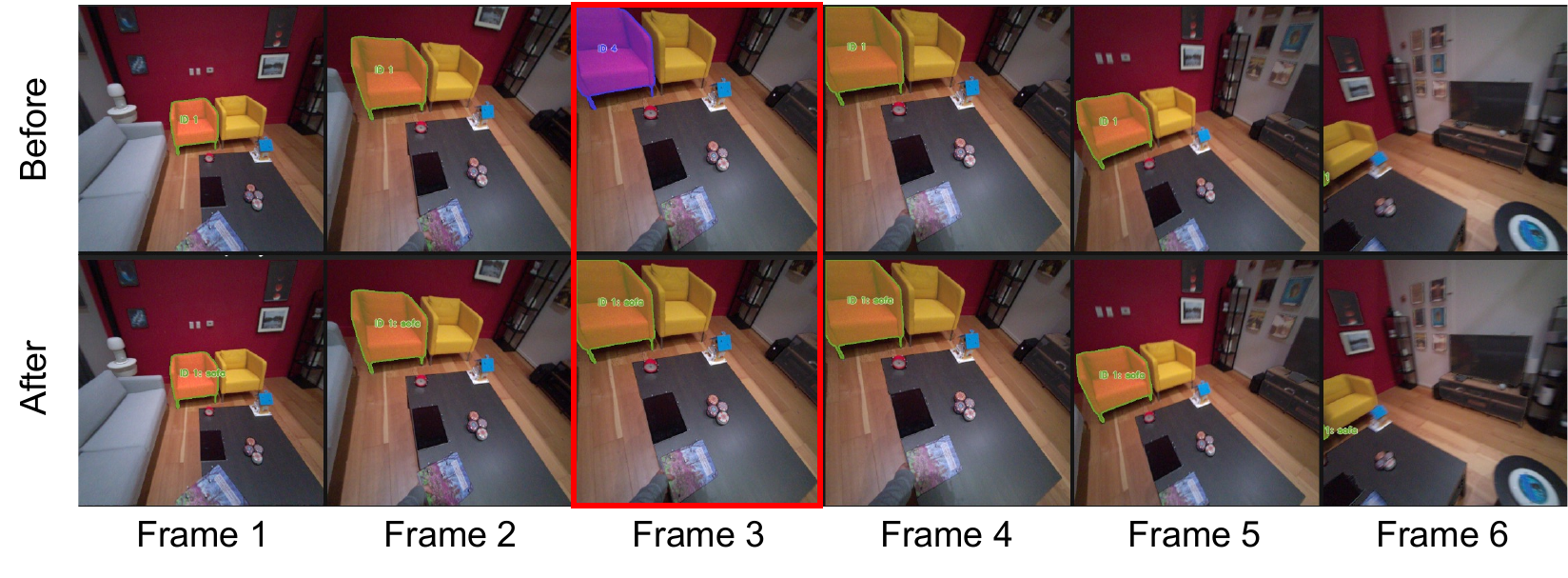}
    \caption{\textbf{Instance-identity inconsistency across frames, and its resolution by fusion.} \textbf{Top row:} the example of per-frame SAM~3 masks from the segmentation stage (Sec.~\mainsec{3.1}); the same seat is assigned \textbf{\textcolor{green}{ID~1: \texttt{sofa}}} in frames 1-2 and 4-6 but a distinct \textbf{\textcolor{violet}{ID~4: \texttt{armchair}}} in frame~3 (denoted by \textcolor{red}{red box}), so its identity is not consistent over the sequence. \textbf{Bottom row:} fusion relabels every frame to one canonical instance, yielding the temporally consistent 3D instance used in the scene graph $\mathcal{G}$. Note that the fused 3D instance projected back into each RGB frame for better visual understanding.}
    \label{fig:fusion_failure}
\end{figure}

%% file: corl26/suppl/figures/4_corr_input.tex
\begin{figure}[h]
    \centering
    \includegraphics[width=1\linewidth]{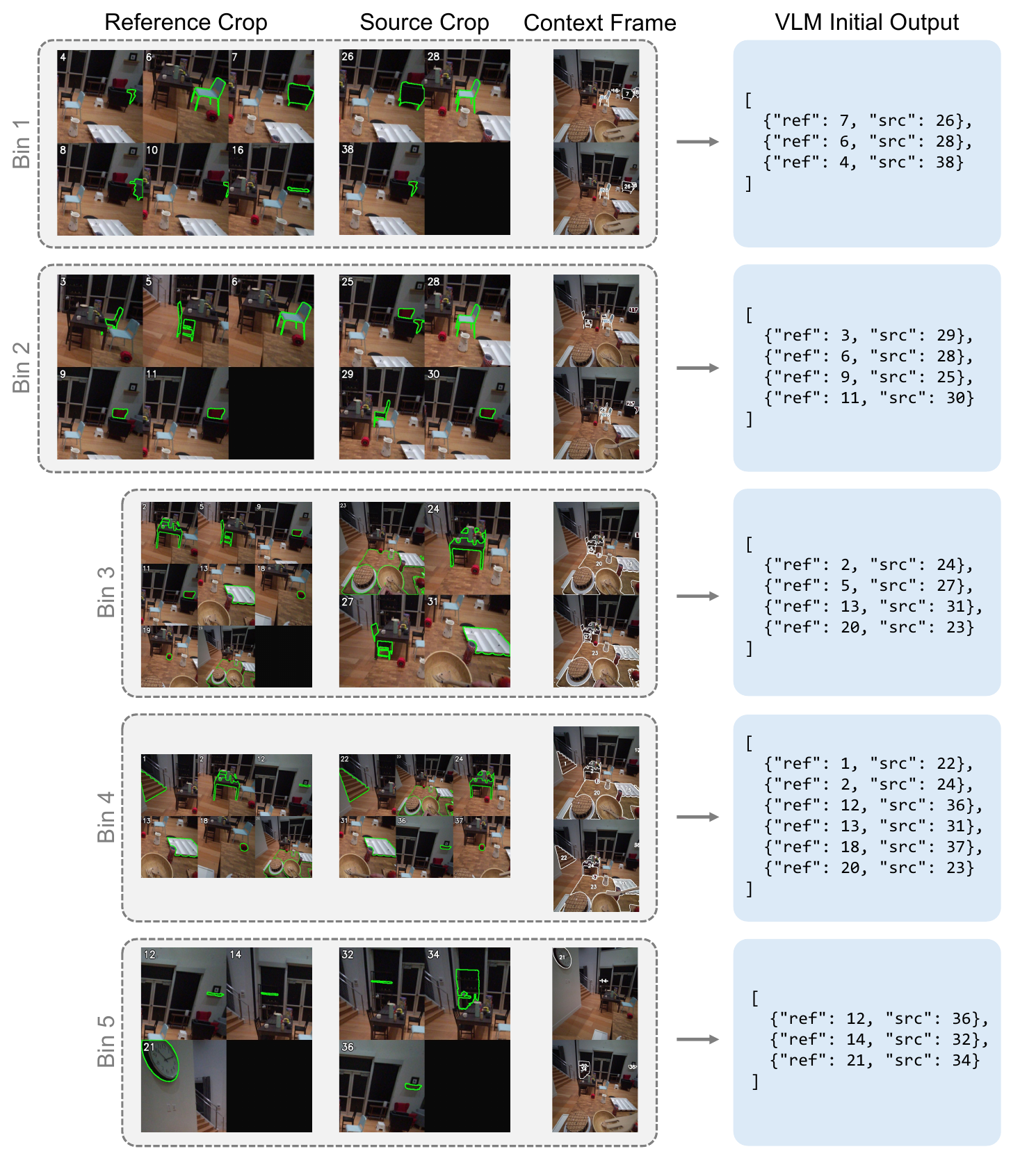}
    \caption{\textbf{Per-bin correspondence query.} For each of the $K{=}5$ height bins (rows), the VLM receives three images (a REF crop grid, a SRC crop grid, and a context frame) in which every instance is outlined in green and stamped with a shared-namespace Set-of-Marks ID (REF markers $1\dots N$, SRC markers $N{+}1\dots N{+}M$, disjoint across the two scans). From these the model returns its initial list of candidate cross-scan pairs (right column), each a \texttt{\{"ref": id, "src": id\}} entry, so that a returned marker pair is directly an instance pair. These candidates are not final: they are filtered by the pairwise verification of Sec.~\ref{sec:suppl-verification} (Fig.~\ref{fig:s4_output}).}
    \label{fig:s4_input}
\end{figure}

%% file: corl26/suppl/tables/correspondence_prompt.tex
\begin{table}[t]
\centering
\caption{\textbf{Per-bin correspondence prompt.} System (top) and user (bottom) sent to the VLM for each bin's crops-plus-context images. The reported configuration runs the non-thinking-mode variant and presents three images (REF crop grid, SRC crop grid, context frame).}
\label{tab:suppl-prompts-som}
\small
\resizebox{0.98\textwidth}{!}{
\begin{tabularx}{\linewidth}{@{}lX@{}}
\toprule
\textbf{Role} & \textbf{Verbatim text} \\
\midrule
SYSTEM & ``You match objects across two captures of the same room. Every object shown here sits within the SAME narrow height band of the room, so they are mutually plausible matches.\par\smallskip You see up to three images: (1) REF crop grid --- a grid of zoomed crops, one per REF object. Each crop is centered on ONE object; its silhouette is outlined in green and its numeric ID is stamped white at the top-left corner. (2) SRC crop grid --- the same, one zoomed crop per SRC object. (3) Context frame --- top = REF, bottom = SRC --- one wide RGB frame per capture showing those objects in their spatial layout, each outlined and labelled with the same IDs.\par\smallskip REF and SRC IDs are disjoint integers. The outline and label are localization aids --- judge the OBJECT itself (color, material, shape, size, and its neighbours in the context frame), never the marker.\par\smallskip Pair a REF ID with a SRC ID only when you are confident they tag the SAME physical object. Most objects appear in only one capture --- leave those unmatched. A wrong match is worse than a missed one.\par\smallskip Output ONLY a JSON array of objects: \texttt{[\{"ref": <id>, "src": <id>\}, ...]}; emit \texttt{[]} if nothing matches. No prose, no fences.'' \\
\addlinespace
USER & ``Match REF and SRC IDs that tag the SAME physical object.\par\smallskip \texttt{REF IDs: [...]}\par \texttt{SRC IDs: [...]}''
\begin{itemize}[leftmargin=1.5em, itemsep=1pt, topsep=2pt]
  \item Use only the IDs listed above.
  \item An ID may appear in at most one pair.
  \item Emit \texttt{[]} if nothing matches.
  \item Output ONLY a JSON array of objects, one per match: \texttt{[\{"ref": <id>, "src": <id>\}, ...]} --- no prose, fences, or preamble.
\end{itemize}
\textit{(The \texttt{REF IDs: [...]} and \texttt{SRC IDs: [...]} lines are filled in per bin with the markers present in that bin.)} \\
\bottomrule
\end{tabularx}
}
\end{table}

%% file: corl26/suppl/tables/verification_prompt.tex
\begin{table}[t]
\centering
\caption{\textbf{Pairwise verification prompts.} The affirmative pair (rows 1--2) implements the \texttt{same?} query of Eq.~(\ref{eq:suppl-verify}); the negated pair (rows 3--4) implements the \texttt{different?} query. A candidate pair survives only when both pairs agree.}
\label{tab:suppl-prompts-verify}
\small
\begin{tabularx}{\linewidth}{@{}lX@{}}
\toprule
\textbf{Prompt} & \textbf{Verbatim text} \\
\midrule
Affirmative SYSTEM & ``You compare two object crops from different captures of the same room. The LEFT crop is from Scene A, the RIGHT crop is from Scene B. Both objects are outlined in green. Decide whether they are the same physical object.'' \\
\addlinespace
Affirmative USER & ``Is the object on the LEFT the same physical object as the object on the RIGHT? Consider shape, color, material, and size. A wrong match is worse than a missed one --- only say yes if confident. Answer with a single integer: [1] for yes, [0] for no.'' \\
\midrule
\addlinespace
Negated SYSTEM & ``You compare two object crops from different captures of the same room. The LEFT crop is from Scene A, the RIGHT crop is from Scene B. Both objects are outlined in green. Decide whether they are DIFFERENT objects.'' \\
\addlinespace
Negated USER & ``Are the object on the LEFT and the object on the RIGHT DIFFERENT physical objects? Consider shape, color, material, and size. A missed mismatch is worse than a false alarm --- say yes if in doubt. Answer with a single integer: [1] for yes (different), [0] for no (same).'' \\
\bottomrule
\end{tabularx}
\end{table}

%% file: corl26/suppl/figures/5_corr_output.tex
\begin{figure}[h]
    \centering
    \includegraphics[width=1\linewidth]{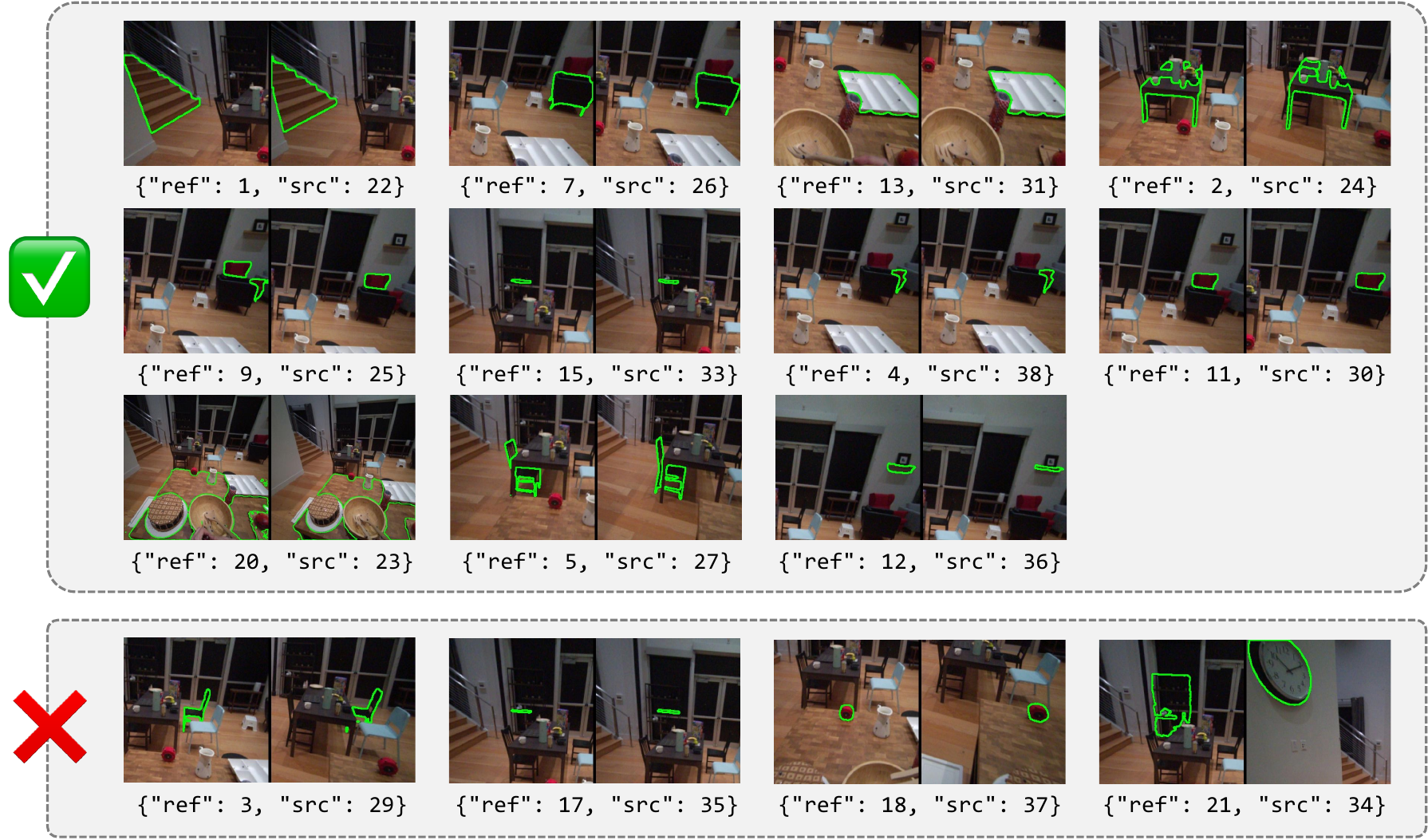}
    \caption{\textbf{Pairwise same/different verification.} Each candidate pair produced by the per-bin query (Fig.~\ref{fig:s4_input}) is rendered as a single side-by-side image of its two crops and passed through the two complementary prompts of Eq.~(\ref{eq:suppl-verify}). Pairs accepted by both prompts (top block, green check) enter the final cross-scan correspondence set $\mathcal{C}$; pairs the gate rejects (bottom block, red cross) are discarded. Each tile is labelled with its \texttt{\{"ref": id, "src": id\}} markers from Fig.~\ref{fig:s4_input}.}
    \label{fig:s4_output}
\end{figure}

%% file: corl26/suppl/3_exp_details.tex
\clearpage
\section{Experimental Setup Details}
\label{sec:suppl-setup-details}
\vspace{-1mm}

This section expands the experimental setup compressed into the main paper's implementation details and experimental setting (Sec.~\ref{sec:exp-impl}, Sec.~\ref{sec:exp-setup}) into the concrete specifics needed to reproduce the reported results: how the evaluation pairs are constructed, how every metric in the main tables is defined, how the baselines are run, and the exact hyperparameters, checkpoints, and environment.

\subsection{Dataset Construction and Subscan Pairing}
\label{sec:suppl-dataset}
\vspace{-1mm}

\method is evaluated on two egocentric benchmarks of diverse indoor scenes, Aria Digital Twin (ADT)~\cite{pan2023aria} and Aria Everyday Activities (AEA)~\cite{lv2024aria}. Both are RGB-only egocentric captures: \method consumes only the head-mounted RGB stream, with no depth sensor and no externally supplied SLAM trajectory.

\textbf{Per-frame inputs.} Each recording is sampled in time at a fixed interval of $0.5$\,s, and every retained RGB frame is rectified from the Aria fisheye stream to a $512$-pixel linear-pinhole image. A short, contiguous run of these sampled frames forms a \emph{subscan}: a partial egocentric observation of the scene at one point in time.

\textbf{Subscan pairs.} The registration task aligns two subscans of the same scene captured at different times, so the unit of evaluation is a \emph{subscan pair}. From the $184$ ADT sequences we derive $6{,}657$ subscan pairs, organized into two regimes: \emph{Single-ARIA}, where the two subscans of a pair come from a single wearer, and \emph{Multi-ARIA}, where they come from multiple wearers of the same scene. This pairing yields the column groups (Single-ARIA / Multi-ARIA / Total) reported on ADT in the main registration result table (Tab.~\ref{tab:adt_aea_main}).
From the $9$ AEA sequences we similarly derive $675$ subscan pairs; because every AEA recording comes from a single wearer, AEA has no Multi-ARIA regime and contributes only single-wearer pairs.

\textbf{Ground-truth availability and the GT/reconstructed split.} The two benchmarks differ in what supervision they provide, which determines the point clouds on which each is evaluated. ADT provides ground-truth point clouds, camera poses, and object lists, so ADT can be evaluated in two settings: on the ground-truth clouds, and on clouds reconstructed from RGB by VGGT-$\Omega$~\cite{wang2026vggt} (the sensor-free setting \method targets). AEA, by contrast, provides only camera poses and a semi-dense point cloud and no dense ground-truth geometry; we therefore evaluate AEA exclusively on VGGT-$\Omega$-predicted clouds. This asymmetry is reflected in Tab.~\ref{tab:adt_aea_main}, where AEA appears only under the reconstructed-cloud setting.

\subsection{Task and Metric Definitions}
\label{sec:suppl-metrics}
\vspace{-1mm}

Here we define the tasks and metrics behind every number in the main result tables; Tab.~\ref{tab:suppl-metrics} consolidates the metric definitions, directions, and thresholds for quick lookup.

\textbf{Tasks.} We evaluate three tasks. \emph{(i) Registration} recovers the rigid transform $T\in SE(3)$ that aligns the two subscans' point clouds of a pair. \emph{(ii) Instance correspondence} matches object instances across the two subscans; its ground-truth pairs are formed by thresholding the intersection-over-union (IoU) of mask-projected instances under point-wise mutual nearest-neighbor matching. \emph{(iii) Object listing} compares a subscan's predicted object list against the ground-truth list, matched via Hungarian assignment on CLIP~\cite{clip} cosine similarity.

\textbf{Registration metrics.} We report four quantities. \emph{Registration Recall} (RR) is the fraction of pairs that are jointly accurate in rotation and translation: a pair counts as recalled only when its Relative Rotation Error is below $5^\circ$ \emph{and} its Relative Translation Error is below $0.2$\,m. \emph{Relative Rotation Error} (RRE) and \emph{Relative Translation Error} (RTE) are the angular and Euclidean errors of the estimated transform against ground truth. \emph{Valid Ratio} (VR) is the fraction of pairs for which the method returns a non-degenerate transform at all; a pair is counted as valid when the estimator yields a usable rigid transform rather than flagging failure (for example failing to converge or producing no transform), and invalid otherwise. VR thus separates outright failures (\eg\ non-convergence) from inaccurate-but-produced estimates, so a method can have high VR yet low RR.

\textbf{Correspondence metrics.} We report \emph{Node Precision} (NP) and \emph{Node Recall} (NR), their \emph{F1}, and the \emph{Coverage Ratio} (CR), defined as the fraction of pairs for which at least one correct instance match is produced. These node metrics are scored against ground-truth correspondences, which we obtain following SG-Reg~\cite{11024207}: instance segmentation masks are back-projected into 3D and matched by an IoU over radius-$r$ neighborhoods in the global frame. Given back-projected point clouds $P_\text{ref}, P_\text{src}$, where $B_r(p_i) \cap P_\text{src}$ is the set of points in $P_\text{src}$ within radius $r$ of $p_i$, the SG-Reg IoU is
\begin{equation}
    \text{IoU}_{\text{SG-Reg}} = \frac{C}{|P_\text{ref}| + |P_\text{src}| - C}, \qquad \text{for } C = \bigl|\{{p_i \in P_\text{ref} \mid B_r(p_i) \cap P_\text{src} \neq \emptyset}\}\bigr|.
\end{equation}

Because SG-Reg assumes scanned or ground-truth clouds, this formula becomes asymmetric under a large density difference and can yield invalid values ($\text{IoU} > 1$); voxelizing mitigates this in some cases but degrades object segmentation on noisy reconstructions. We therefore use a symmetric mutual-nearest-neighbor IoU that is agnostic to point-cloud density and preserves mask quality:
\begin{equation}
    \text{IoU}_{\text{sym}} = \frac{m_\text{ref, src} + m_\text{src, ref}}{|P_\text{ref}| + |P_\text{src}|}, \qquad \text{for } m_{X,Y} = \bigl|\{{p_i \in P_X \mid \min_{q \in P_Y}|p_i - q| \leq r}\}\bigr|.
\end{equation}

The fixed denominator keeps $\text{IoU}_{\text{sym}} \in [0, 1]$, and the metric is symmetric by construction: swapping $P_\text{ref} \leftrightarrow P_\text{src}$ leaves the value unchanged. A pair of instances is a ground-truth match when its $\text{IoU}_{\text{sym}}$ exceeds $\tau_{\text{IoU}}{=}0.3$ at radius $r{=}0.2$\,m.

\textbf{Object-listing metrics.} We report Precision (P), Recall (R), F1, and Anchor Recall (AR), all computed at a CLIP-similarity threshold of $0.5$ (denoted @50), \ie, a predicted label is credited to a ground-truth label only when their CLIP cosine similarity exceeds $0.5$. Precision, recall, and F1 are taken over all ground-truth objects; anchor recall is taken over the subset of ground-truth objects \emph{shared across both subscans} of a pair, which are salient landmarks that make up the best correspondence targets.

\begin{table}[t]
\centering
\small
\caption{\textbf{Reference card for the metrics defined in this subsection.} ($\uparrow$/$\downarrow$: higher/lower is better)}
\label{tab:suppl-metrics}
\resizebox{0.98\textwidth}{!}{
\begin{tabular}{l c l l}
\toprule
\textbf{Metric} & \textbf{Dir.} & \textbf{Definition} & \textbf{Threshold / note} \\
\midrule
RR & $\uparrow$ & Registration Recall & RRE $<5^\circ$ and RTE $<0.2$\,m \\
RRE & $\downarrow$ & Relative Rotation Error & degrees \\
RTE & $\downarrow$ & Relative Translation Error & meters \\
VR & $\uparrow$ & Valid Ratio & non-degenerate $T$ returned \\
NP / NR / F1 & $\uparrow$ & Node Precision / Recall / F1 & $\tau_{\text{IoU}}{=}0.3$, $r{=}0.2$\,m; macro over pairs \\
CR & $\uparrow$ & Coverage Ratio & $\ge 1$ correct match per pair \\
P / R / F1 (listing) & $\uparrow$ & Precision / Recall / F1 of object names & CLIP-sim $\ge 0.5$ (@50) \\
AR (listing) & $\uparrow$ & Anchor Recall over shared GT objects & CLIP-sim $\ge 0.5$ (@50) \\
\bottomrule
\end{tabular}
}
\end{table}

\subsection{Baseline Protocols}
\label{sec:suppl-baselines}

\textbf{Registration baselines.} We compare against two families. The \emph{scene-level} family registers the full scene point cloud directly and comprises TEASER++~\cite{yang2020teaser}, a certifiable global registration method run with two descriptor variants for its correspondences (handcrafted FPFH and learned FCGF); GeoTransformer~\cite{qin2022geometric}, a learned coarse-to-fine point-cloud matcher; and BUFFER-X~\cite{Seo_BUFFERX_arXiv_2025}, a generalizable point-cloud registration pipeline. The \emph{scene-graph} family is our main comparison, SG-Reg~\cite{11024207}, trained on curated indoor RGB-D scene graphs.

Every registration baseline is run from the authors' officially released implementation at its recommended default indoor configuration, with no per-dataset re-tuning of the descriptor, matcher, or robust-estimator settings, and the same configuration is used across the GT and reconstructed-cloud settings.

Under this protocol, two baselines appear as non-numeric cells in the registration table (Tab.~\ref{tab:adt_aea_main}): TEASER++ with FCGF descriptors fails to converge on the reconstructed clouds (while still running on the ground-truth clouds), and GeoTransformer exhausts the single-GPU memory budget at the scene level in both settings. Both are reported as observed at the methods' released configurations.

\textbf{Correspondence baselines.} For the correspondence comparison (Tab.~\ref{tab:correspondence}), the open-vocabulary detection-and-embedding baseline matches instances purely by visual-feature similarity, in contrast to \method's VLM reasoning over shared-namespace Set-of-Marks prompts (Sec.~\ref{sec:suppl-binning}). For each instance, we select a representative frame, the frame in which that instance's segmentation mask has the largest area, and crop the instance's bounding box from it with a $10\%$ padding margin. Each crop is embedded by concatenating two backends: the Grounding DINO~\cite{liu2024grounding} detector's decoder feature and the CLIP~\cite{clip} ViT-L/14 image embedding, and the concatenated vector is L2-normalized. Cross-scan instances are then matched by Hungarian assignment on the cosine-similarity matrix between the two subscans' instance embeddings, keeping an assigned pair only when its cosine similarity is at least $0.5$. SG-Reg~\cite{11024207} also appears in the correspondence comparison, reusing the instance matches produced by its learned scene-graph matcher.

\subsection{Hyperparameters, Checkpoints, and Environment}
\label{sec:suppl-hparams}

\textbf{Models and checkpoints.} \method\ uses three pretrained foundation models off the shelf, without any fine-tuning: VGGT-$\Omega$~\cite{wang2026vggt} as the geometric backbone that predicts per-frame depth, intrinsics, and pose from RGB; Qwen3.6-27B~\cite{qwen3.6-27b} as the vision-language model for object listing and cross-scan instance matching; and SAM~3~\cite{carion2025sam} for text-prompted video segmentation. For the final per-instance point-level matching we report three interchangeable descriptors, all used as released: FCGF~\cite{FCGF2019} and GeoTransformer~\cite{qin2022geometric} from the authors' 3DMatch-trained checkpoints, and the handcrafted FPFH~\cite{rusu2009fast}, which has no learned weights. Their head-to-head comparison is reported with the registration results.

\textbf{Pipeline hyperparameters.} The following numeric settings are used unchanged for every reported \method\ result. In per-instance fusion, the pooled cloud is voxelized at $5$\,cm, each voxel is relabeled to its majority instance ID, and an instance is absorbed into another when at least $50\%$ of its points are reallocated. In correspondence, instances are split into $K{=}5$ equal-count quantile height bins with $20\%$ overlap between adjacent bins. In registration, RANSAC is run for $10$K iterations with a $3$\,cm inlier threshold. All runs use a fixed random seed of $42$ and execute on a single H200 GPU.

%% file: corl26/suppl/4_additional_quan.tex
\section{Additional Quantitative Results}
\label{sec:suppl-addtional-quan}

\subsection{VLM Backbone Ablation}

\providecommand{\best}[1]{\cellcolor{orange!20}{\textbf{#1}}}
\providecommand{\second}[1]{\cellcolor{orange!10}{\underline{#1}}}

\begin{wraptable}{r}{0.62\textwidth}
\centering
\vspace{-4mm}
\caption{\textbf{VLM backbone ablation for scene registration.} The \colorbox{orange!20}{\textbf{best}} and \colorbox{orange!10}{\underline{second best}} per column are highlighted.}
\vspace{-1mm}
\label{tab:vlm_ablation_adt}
\setlength{\tabcolsep}{6pt}
\resizebox{0.6\textwidth}{!}{%
\begin{tabular}{l cccc}
\toprule
\textbf{Model}
& \makecell{RR $\uparrow$ \\ (\%)} & \makecell{RRE $\downarrow$ \\ ($^{\circ}$)}
& \makecell{RTE $\downarrow$ \\ (m)} & \makecell{VR $\uparrow$ \\ (\%)} \\
\midrule
\multicolumn{5}{c}{\cellcolor{gray!8}\textit{Open-source (open-weight)}} \\
\midrule
Qwen3.6-27B~\cite{qwen3.6-27b}    & \best{82.7}   & \best{5.00}   & \best{0.24}   & \best{99.4}   \\
Qwen3-VL-32B-Instruct~\cite{bai2025qwen3}        & 67.4          & 13.83         & 0.63          & 92.5          \\
InternVL3-14B~\cite{zhu2025internvl3}                & 55.4          & 23.17         & 1.04          & 84.1          \\
InternVL3-8B~\cite{zhu2025internvl3}                 & 34.6          & 35.95         & 1.70          & 70.1          \\
\midrule
\multicolumn{5}{c}{\cellcolor{gray!8}\textit{Closed-source (API-only)}} \\
\midrule
GPT-5.4~\cite{openai2026gpt54}                      & \second{74.0} & \second{8.32} & \second{0.43} & \second{94.3} \\
GPT-4o~\cite{hurst2024gpt}                       & 51.5          & 23.99         & 1.19          & 87.9          \\
\bottomrule
\end{tabular}%
}
\vspace{-12pt}
\end{wraptable}

\method delegates two stages to a VLM: open-vocabulary object listing and cross-scan instance correspondence. To quantify how much this choice matters, we swap the VLM backbone while holding every other stage fixed and re-evaluate registration. All backbones run on the same point clouds reconstructed from RGB by VGGT-$\Omega$~\cite{wang2026vggt}, and share GeoTransformer~\cite{qin2022geometric} as the final registration descriptor, so any difference is attributable to the VLM alone. Querying the closed, API-served models over the full benchmark is cost-prohibitive, so this ablation is run on a smaller common subset, the ADT Single-ARIA \emph{meal} split ($16$ sequences, $561$ subscan pairs), with every backbone evaluated on that same subset. 

Table~\ref{tab:vlm_ablation_adt} reports the outcome. Registration quality tracks VLM capability rather than model identity: our default backbone, Qwen3.6-27B~\cite{qwen3.6-27b}, leads on all four metrics, and the strongest API model, GPT-5.4~\cite{openai2026gpt54}, is second on each. Capability matters more than raw parameter count, as Qwen3.6-27B surpasses the larger Qwen3-VL-32B-Instruct~\cite{bai2025qwen3}. The weakest backbones, InternVL3-8B~\cite{zhu2025internvl3} and GPT-4o~\cite{hurst2024gpt}, fall off sharply on every metric, including Valid Ratio (VR), so a weaker VLM yields not only less accurate transforms but fewer usable ones.

More importantly, because \method uses the VLM off the shelf without any training, this trend is a strength rather than a liability: as stronger VLMs are released, the pipeline can adopt them directly and improve with no retraining or architectural change, so its accuracy is positioned to rise with future progress in VLMs at no additional cost.

\begin{table}[h]
\centering
\setlength{\tabcolsep}{5pt}
\captionsetup{width=0.9\textwidth}
\caption{\textbf{Quantitative comparison between open-vocabulary taggers and VLM.} Each lister is run with every other stage fixed. Evaluated on ADT dataset.}
\label{tab:object_listing}
\resizebox{.9\textwidth}{!}{%
\begin{tabular}{l cccc cccc}
\toprule
& \multicolumn{4}{c}{\textbf{Object listing}} & \multicolumn{4}{c}{\textbf{Registration}} \\
\cmidrule(lr){2-5} \cmidrule(lr){6-9}
\textbf{Method} & AR@50 $\uparrow$ & P@50 $\uparrow$ & R@50 $\uparrow$ & F1@50 $\uparrow$ & RR $\uparrow$ & RRE $\downarrow$ & RTE $\downarrow$ & VR $\uparrow$ \\
\midrule
RAM \citep{zhang2024recognize}            & \best{99.2} & 72.0 & \best{98.2} & 81.0 & \second{55.0} & \second{13.54} & \second{0.60} & \second{94.1} \\
RAM++ \citep{huang2025open}      & \best{99.2} & 68.7 & \second{97.9} & 78.8 & 54.0 & 14.50 & 0.63 & 93.6 \\
Tag2Text \citep{huang2024tag2text}       & 90.1 & \second{89.7} & 86.0 & \second{85.1} & 51.5 & 16.84 & 0.72 & 92.4 \\
\midrule
\textbf{VLM (Qwen3.6-27B~\cite{qwen3.6-27b})} & \second{92.4} & \best{95.8} & 87.0 & \best{90.0} & \best{56.2} & \best{12.55} & \best{0.59} & \best{96.0} \\
\bottomrule
\end{tabular}}
\end{table}

\subsection{Object Listing Ablation}

\method builds its object list by prompting the VLM to list the objects in a subscan, an open-vocabulary alternative to dedicated image taggers. To assess this choice, we replace the object lister with three open-vocabulary taggers, RAM~\citep{zhang2024recognize}, RAM++~\citep{huang2025open}, and Tag2Text~\citep{huang2024tag2text}, holding the rest of the pipeline fixed, including the VGGT-$\Omega$ clouds and the GeoTransformer registration backend. Table~\ref{tab:object_listing} reports both the semantic object listing quality and the resulting registration on the reconstructed ADT clouds; the VLM row corresponds to our reported configuration (Tab.~\ref{tab:adt_aea_main}).

On listing quality, \method is the most precise: it leads on Precision ($95.8$) and F1 ($90.0$). The taggers instead saturate recall, with RAM and RAM++ reaching $98.2$ and $97.9$ Recall and $99.2$ Anchor Recall but only $72.0$ and $68.7$ Precision, as they emit many labels, including objects that are not present. Tag2Text is more balanced but still trails \method on Precision and F1.

Crucially, this carries through to registration. With every other stage held fixed, \method's lister yields the best registration on all four metrics, ahead of every tagger. The taggers' near-saturated listing recall does not help; their low precision hurts, since spurious object names become spurious instances that degrade correspondence and pose estimation. Precise object listing, not exhaustive tagging, is what drives downstream registration, which is why \method uses a VLM rather than an off-the-shelf tagger.

\subsection{Segmentation and Fusion Analysis}
\label{sec:supp_segfusion}

\begin{wrapfigure}{r}{.28\linewidth}
    \vspace{-4mm}
    \centering
    \includegraphics[width=\linewidth]{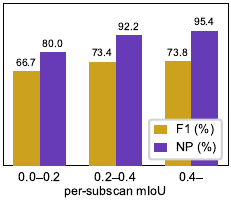}
    \caption{Node matching vs.\ post-fusion mask mIoU.}
    \label{fig:segfusion}
    \vspace{-3mm}
\end{wrapfigure}
Correspondence and pose estimation depend on the instance masks produced by SAM~3~\cite{carion2025sam} and consolidated by per-instance fusion (Sec.~\ref{sec:suppl-fusion}). We assess this dependency on the ADT Single-ARIA meal split.

\textbf{Segmentation quality.} We bin subscans by the mIoU of their post-fusion instance masks against ADT ground truth and report correspondence quality per bin (Fig.~\ref{fig:segfusion}). Node matching degrades gracefully as segmentation quality decreases. Registration is less sensitive than correspondence, since inlier-ratio voting requires a single correct pair rather than many, so coverage suffices.

\newpage
\WFclear
\begin{wraptable}{r}{.60\linewidth}
    \centering
    \caption{Sensitivity to fusion hyperparameters.}
    \label{tab:fusion_params}
    \small
    \begin{tabular}{lccc}
        \toprule
        Setting & NP $\uparrow$ & CR $\uparrow$ & RR $\uparrow$ \\
        \midrule
        \rowcolor{gray!18}
        \textbf{Ours (5\,cm, 50\%)} & 94.0 & 60.2 & 82.7 \\
        Voxel: 3\,cm / 8\,cm & 93.9 / 92.7 & 61.1 / 60.5 & 81.2 / 80.4 \\
        Absorb: 30\% / 70\% & 92.5 / 93.6 & 59.6 / 62.5 & 82.9 / 80.6 \\
        \bottomrule
    \end{tabular}
    \vspace{-3mm}
\end{wraptable}
\textbf{Fusion hyperparameters.} Table~\ref{tab:fusion_params} varies the voxel size and absorption threshold around our setting (5\,cm, 50\%). NP and RR vary by at most $1.2$ and $2.3$ points, respectively, and the near-duplicate merge rarely fires at this operating point, so fusion is insensitive to these choices within the tested range.

\subsection{Inference Time}

\begin{wraptable}{r}{0.6\textwidth}
\centering
\vspace{-\intextsep}
\caption{\textbf{End-to-end inference time of \method.} Each entry is the time a stage contributes to one registration.}
\vspace{-2mm}
\label{tab:suppl-timing}
\small
\setlength{\tabcolsep}{5pt}
\resizebox{\linewidth}{!}{%
\begin{tabular}{l l r}
\toprule
\textbf{Stage} & \textbf{Model / component} & \textbf{Time} \\
\midrule
Per-frame geometry        & VGGT-$\Omega$~\cite{wang2026vggt}  & $1.06$\,s \\
Object listing            & Qwen3.6-27B~\cite{qwen3.6-27b}    & $4.48$\,s \\
Instance segmentation     & SAM~3~\cite{carion2025sam}          & $13.84$\,s \\
Per-instance fusion       & geometric      & $0.18$\,s \\
Cross-scan correspondence & Qwen3.6-27B~\cite{qwen3.6-27b}    & $22.34$\,s \\
Registration              & GeoTransformer~\cite{qin2022geometric} & $7.47$\,s \\
\midrule
\textbf{End-to-end} & & \textbf{49.37\,s} \\
\bottomrule
\end{tabular}}
\vspace{-2mm}
\end{wraptable}

Table~\ref{tab:suppl-timing} reports the wall-clock time of each stage of \method for one subscan-pair registration, measured on the single H200 GPU used for all our experiments with GeoTransformer as the registration descriptor.

Per registration, PROSE issues $34.7$ VLM calls on average: $2$ for object listing, one per subscan; $6$ for cross-scan correspondence, one per height bin plus the cross-bin pass; and $2$ per verification candidate, for the affirmative and negated queries. 
End to end, one registration takes approximately $49.37$\,s, dominated by the cross-scan correspondence query ($22.34$\,s) and instance segmentation ($13.84$\,s), while the geometric stages are comparatively light. 

\method targets \emph{offline} registration, aligning two scans of a scene captured at different times rather than running in a real-time loop, so this per-pair latency is acceptable in that setting rather than a real-time bottleneck. The cost also sits almost entirely in off-the-shelf foundation models, so it shrinks directly with faster VLMs and standard serving optimizations, at no change to the method.

%% file: corl26/suppl/5_additional_qual.tex
\section{Qualitative Results}
\label{sec:suppl-addtional-qual}

We visualize \method's quality in terms of registration, correspondence, and scene graph generation in Figs.~\ref{fig:suppl_qual_reg},~\ref{fig:suppl_qual_corr}, and~\ref{fig:suppl_qual_sg}, respectively. 

In Fig.~\ref{fig:suppl_qual_reg}, \method produces visibly superior pose estimates compared to registration baselines on the same subscan pair. Because the point clouds are reconstructed from egocentric RGB, their geometry is noisy and degrades the dense local features that feature-based baselines rely on; \method instead anchors on object-level instance correspondences, which remain stable under this noise.

Figure~\ref{fig:suppl_qual_corr} shows predicted instance pairings across partially overlapping subscans captured from varying viewpoints. The results demonstrate robustness to large viewpoint changes and occasional erroneous matches: as long as a subset of salient objects is correctly paired, registration remains successful. 

Finally, Fig.~\ref{fig:suppl_qual_sg} visualizes the scene graph for the environment used for the robot path planning experiment in Sec.~\ref{sec:exp-pathplanning} (Sec.~\ref{sec:supp-pathplanning} in appendix). Our method produces a large number of instances and captures their semantics through textual labels and spatial locations. Although individual objects may be misplaced or mislabeled, the majority of nodes have realistic centroid placement and category labels, enabling spatial relationship extraction and downstream use.

\input{corl26/suppl/figures/6_qual_reg}

\input{corl26/suppl/figures/7_qual_corr}

\input{corl26/suppl/figures/8_qual_sg}

%% file: corl26/suppl/figures/6_qual_reg.tex
\begin{figure}[h]
    \centering
    \includegraphics[width=1\linewidth]{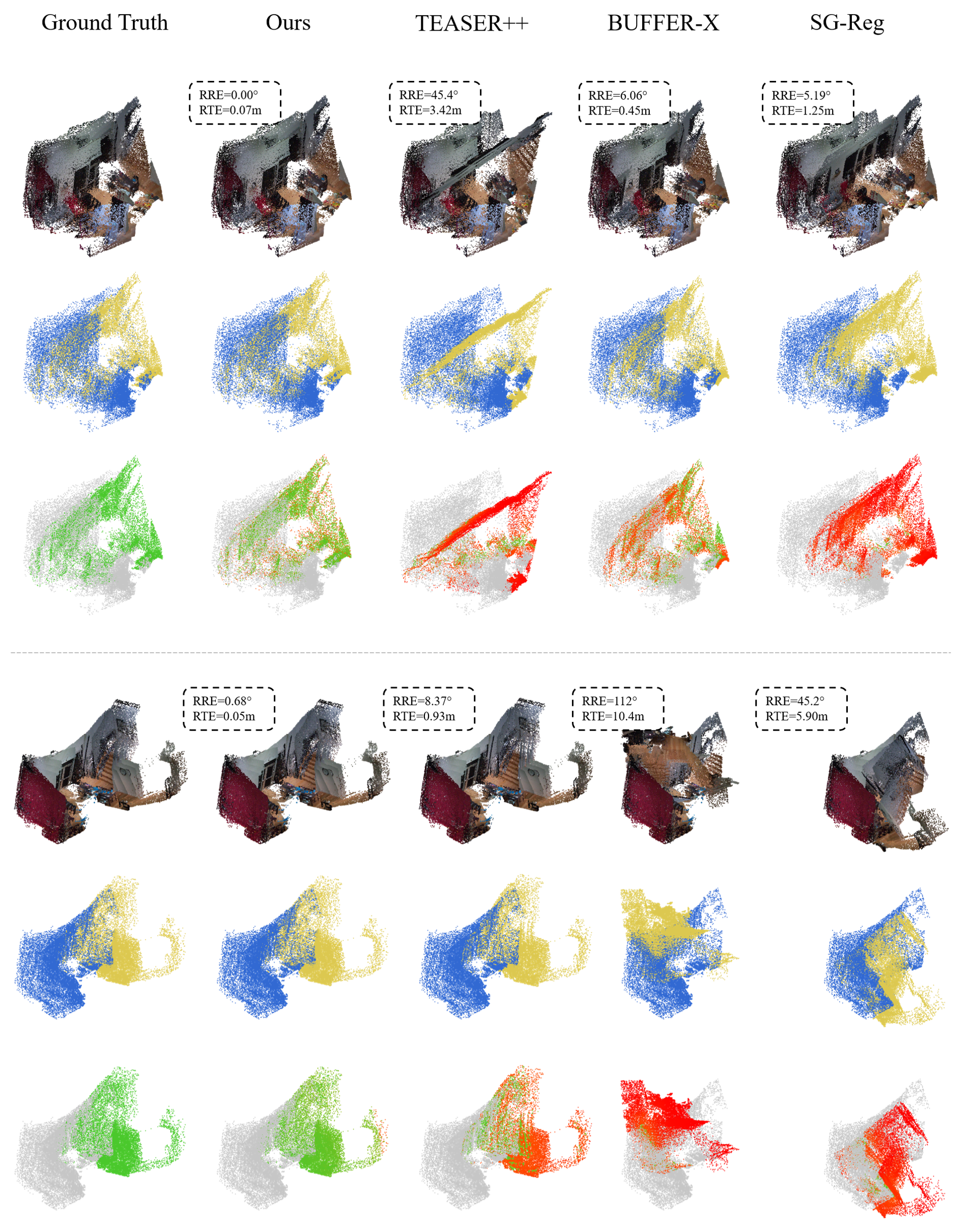}
    \caption{\textbf{Qualitative registration results.} Additional comparisons between registration baselines on the clean ADT sequences. We display RGB reconstruction from VGGT-$\Omega$ (top), colored subscans (middle), and a per-point error heatmap (bottom). The error is the distance to the nearest GT point after registration, where \textcolor{green}{green} indicates low error and \textcolor{red}{red} indicates high error.}
    \label{fig:suppl_qual_reg}
\end{figure}

%% file: corl26/suppl/figures/7_qual_corr.tex
\begin{figure}[h]
    \centering
    \includegraphics[width=1\linewidth]{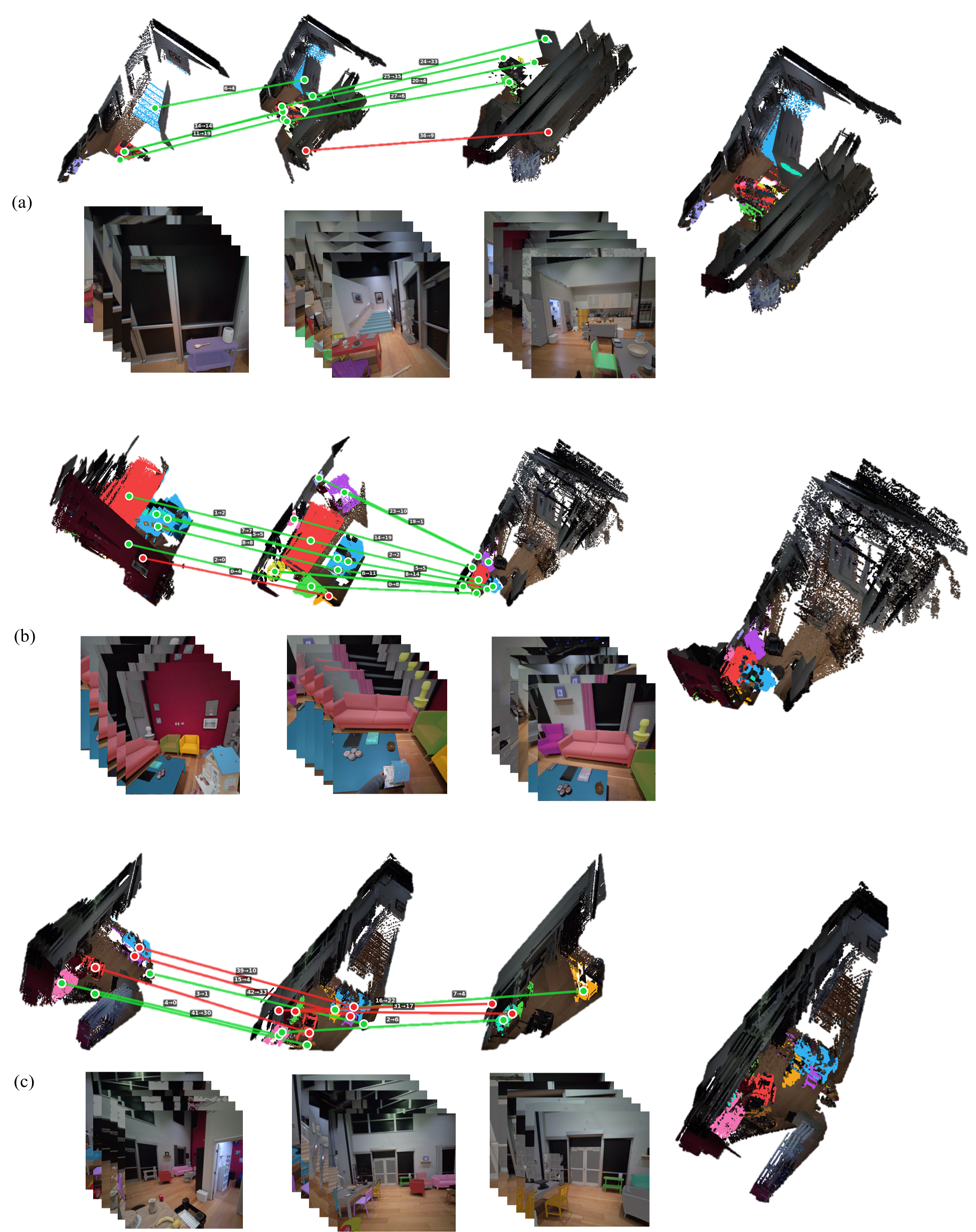}
    \vspace{3mm}
    \caption{\textbf{Qualitative results on correspondence.} Additional correspondence visualizations of \method on three subscans with instance masked frames and subsequent registration. In (a) and (b) frames are captured from differing views, yet we find multiple correct instance matches leading to successful registration. Likewise in (c), our method anchors on a few correct matches and retrieves the transformation despite many erroneous pairs.}
    \label{fig:suppl_qual_corr}
\end{figure}


%% file: corl26/suppl/figures/8_qual_sg.tex
\begin{figure}[h]
    \centering
    \includegraphics[width=1\linewidth]{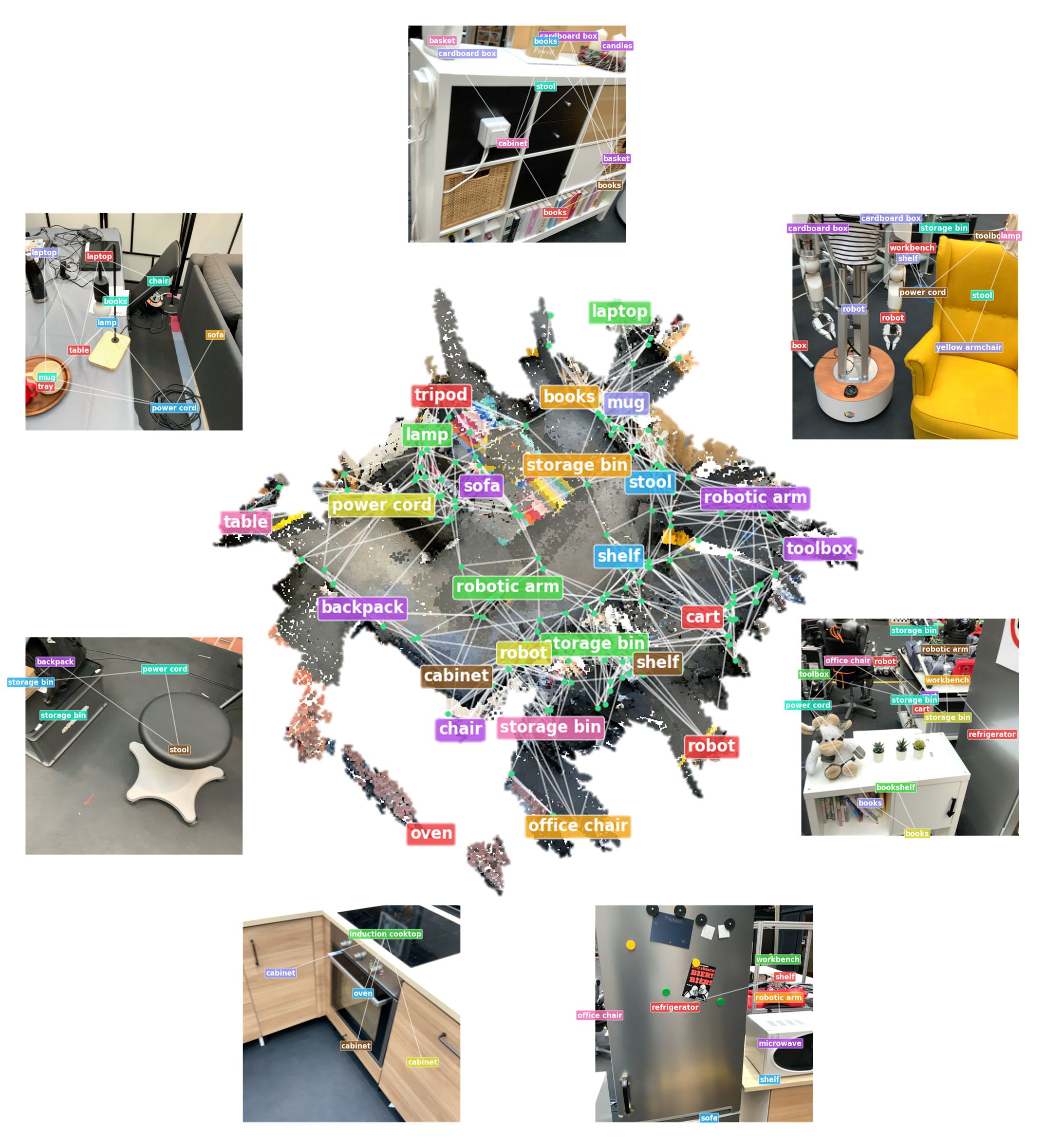}
    \caption{\textbf{Qualitative result on scene graph generation.} Additional scene graph visualization of \method on the path planning environment with nodes and edges overlayed on the VGGT-$\Omega$ reconstruction. We furthest sample nodes to highlight their labels and display individual frames from each section of the room with every node in frame labeled and connected with edges. 
    }
    \label{fig:suppl_qual_sg}
\end{figure}